\documentclass[lettersize,journal]{IEEEtran}
\usepackage{amsmath,amsfonts}
\usepackage{array}
\usepackage[caption=false,font=normalsize,labelfont=sf,textfont=sf]{subfig}
\usepackage{textcomp}
\usepackage{stfloats}
\usepackage{url}
\usepackage{verbatim}
\usepackage{graphicx}
\usepackage{cite}
\usepackage{booktabs}
\usepackage{multirow}
\usepackage{makecell}
\usepackage{graphicx}
\usepackage[linesnumbered,ruled,vlined]{algorithm2e}
\usepackage{algpseudocode}
\usepackage[table,dvipsnames]{xcolor}
\usepackage{amsmath}
\usepackage{bbding}
\usepackage{colortbl}
\definecolor{iccvblue}{rgb}{0.21,0.49,0.74}
\usepackage[pagebackref,breaklinks,colorlinks,allcolors=iccvblue]{hyperref}
\usepackage{cleveref}

\begin{document}

\title{BGM: Background Mixup for X-ray Prohibited Items Detection}

\author{Weizhe Liu, Renshuai Tao, Hongguang Zhu, Yunda Sun, \\Yao Zhao,~\IEEEmembership{Fellow,~IEEE}, Yunchao Wei,~\IEEEmembership{Member,~IEEE}
\thanks{Corresponding author: Yao Zhao.}
\thanks{Weizhe Liu, Renshuai Tao, Yao Zhao and Yunchao Wei are with the Institute of Information Science, Beijing Jiaotong University~(email: \{liuweizhe,rstao,yzhao\}@bjtu.edu.cn, wychao1987@gmail.com).}
\thanks{Hongguang Zhu is with the Faculty of Data Science, City University of Macau~(email: zhuhongguang1103@gmail.com).}
\thanks{Yunda Sun is with the Nuctech Company Limited~(email: sunyunda@nuctech.com).}
}

\markboth{Journal of \LaTeX\ Class Files,~Vol.~14, No.~8, August~2021}%
{Shell \MakeLowercase{\textit{et al.}}: A Sample Article Using IEEEtran.cls for IEEE Journals}


\maketitle

\begin{abstract}
Current data-driven approaches for X-ray prohibited items detection remain under-explored, particularly in the design of effective data augmentations.
Existing natural image augmentations for reflected light imaging neglect the data characteristics of X-ray security images.
Moreover, prior X-ray augmentation methods have predominantly focused on foreground prohibited items, overlooking informative background cues.
In this paper, we propose \textbf{Background Mixup (BGM)}, a background-based augmentation technique tailored for X-ray security imaging domain.
Unlike conventional methods, BGM is founded on an in-depth analysis of physical properties including:
1)~\textbf{X-ray Transmission Imagery}: 
Transmitted X-ray pixels represent composite information from multiple materials along the imaging path.
2)~\textbf{Material-based Pseudo-coloring}: Pseudo-coloring in X-ray images correlates directly with material properties, aiding in material distinction.
Building upon the above insights, BGM mixes background patches across regions on both 1) texture structure and 2) material variation, to benefit models from complicated background cues.
This enhances the model’s capability to handle domain-specific challenges such as occlusion-induced discriminative imbalance.
Importantly, BGM is orthogonal and fully compatible with existing foreground-focused augmentation techniques, enabling joint use to further enhance detection performance.
Extensive experiments on multiple X-ray security benchmarks show that BGM consistently surpasses strong baselines, without additional annotations or significant training overhead.
This work pioneers the exploration of background-aware augmentation in X-ray prohibited items detection and provides a lightweight, plug-and-play solution with broad applicability.
The code will be available at \href{https://github.com/WiZard-Leo/BackgroundMixup}{https://github.com/WiZard-Leo/BackgroundMixup}.
\end{abstract}    

\section{Introduction}
\label{sec:intro}

\IEEEPARstart{T}{errorist} attacks pose a significant threat to public security. 
In the context of counter-terrorism and anti-explosive measures, security inspections play a crucial role in mitigating these risks. 
As a reliable and non-intrusive detection method, X-ray imaging has seen widespread application, detecting explosives, firearms, flammable liquids, and other hidden hazards in luggage.
Meanwhile, with the rapid advancements of computer vision and machine learning~\cite{lecun2015deep,he2016deep,ren2016faster,lin2017focal,cai2018cascade,zhang2022dino, zhu2020deformable, carion2020end, wei2017object, duan2019centernet, lin2017feature, liu2021swin, xu2023comprehensive, liu2024grounding, cheng2024yolo}, numerous researchers are dedicated to developing advanced automated frameworks for precise detection of kinds of prohibited items~\cite{akcay2022towards,isaac2023seeing,gaus2024performance,liu2022data,mery2017logarithmic,schwaninger2007adaptive,wang2021towards,zhao2022detecting,duan2023rwsc,ma2024coarse,velayudhan2022recent,yang2024dual,ma2024towards,wang2025i2ol,li2024detr,akcay2018using}.

\begin{figure}[t]
  \centering
   \includegraphics[width=1.0\linewidth]{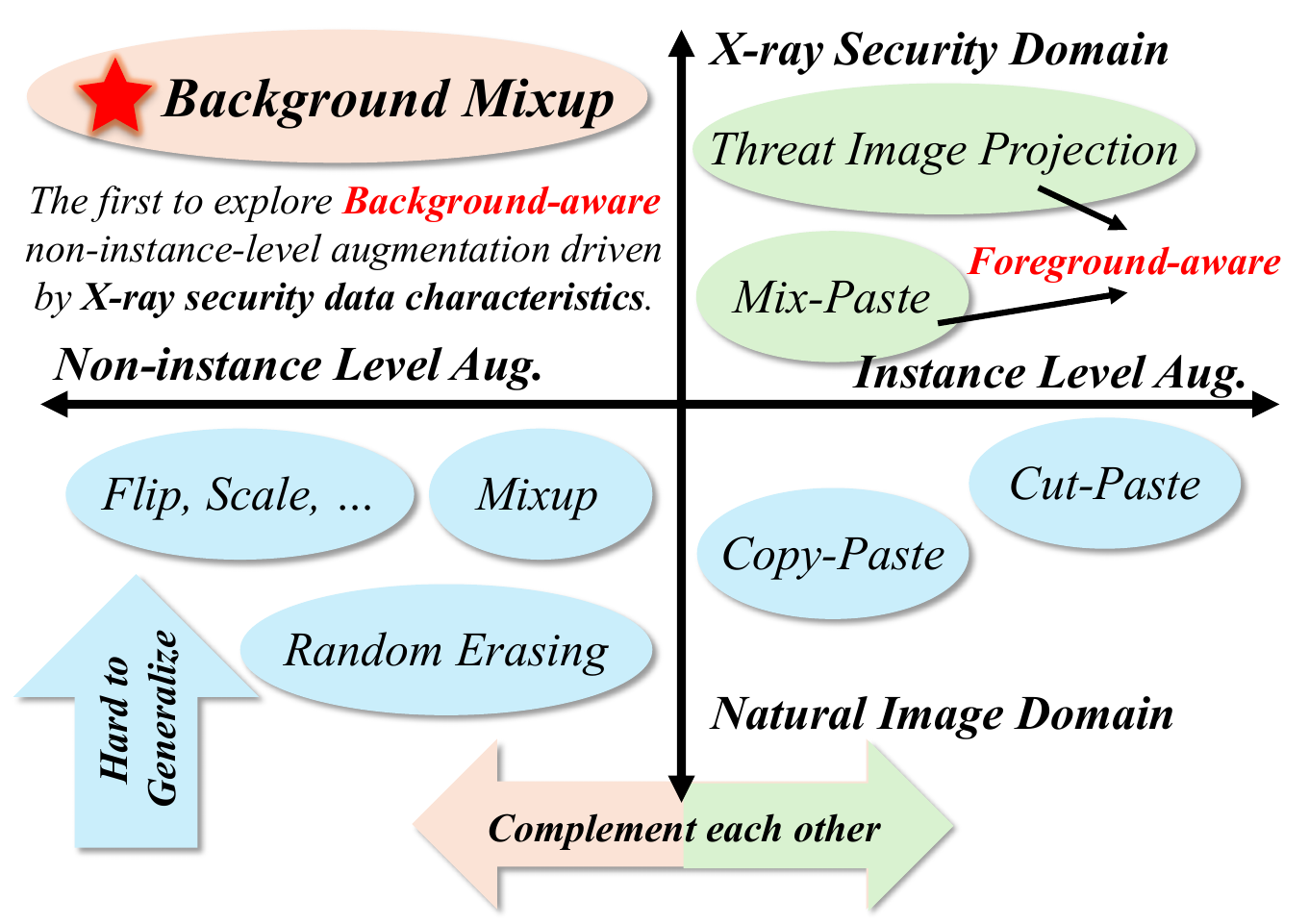}
   \caption{\textbf{Augmentation methods over two physical imaging domains.} 
   \textbf{Firstly}, classical data augmentation methods, primarily designed for the natural image domain, often struggle to generalize effectively to the X-ray security domain. \textbf{Secondly}, existing augmentation methods in the X-ray security domain predominantly focus on instance-level augmentation, overlooking the potential benefits of incorporating background cues. \textbf{Consequently}, we are the first to explore X-ray security data-driven augmentation from a background perspective. Furthermore, background-aware and foreground-aware regularization methods complement each other, leading to improved performance.
   }
   \label{fig:IntroBGM}   
\end{figure}

Despite the abundance of natural image datasets, X-ray security image datasets are significantly smaller due to privacy policies and limited access to X-ray devices. 
In addition, annotating prohibited items requires experienced experts familiar with the nuances of X-ray transmission imaging. Since pixels represent overlapping items rather than single objects as in natural images~\cite{velayudhan2022recent}, it is difficult for untrained personnel to identify prohibited items in the midst of high occlusion and clutter~\cite{akcay2022towards}.
Meanwhile, it is also hard to directly transfer pretrained models on natural images to X-ray security domain, because X-ray images exhibit inherent material-based color variations, insufficient texture information and special occlusion.
\IEEEpubidadjcol
Data-centric learning has propelled substantial advancements in natural image recognition~\cite{lecun2015deep,he2016deep, ren2016faster,cai2018cascade,lin2017focal,dosovitskiy2020image,khan2022transformers,liu2021swin,zhu2020deformable,feng2021tood,zhang2022dino}, supported by large-scale datasets and data augmentation techniques like Random Erasing~\cite{zhong2020random}, Mixup~\cite{zhang2017mixup}, CutMix~\cite{yun2019cutmix} and so on~\cite{dwibedi2017cut,li2021cutpaste,zhang2023pha,ahn2024data,he2024frequency,vaish2024fourier,tripathi2023edges,ghiasi2021simple,kim2020puzzle}. However, as illustrated in Table~\ref{tab:augmentation_comparison}, classical augmentations in natural image domain often fail to boost the detection performance of prohibited items, consistent with previous findings~\cite{webb2021operationalizing}.
For instance, Mixup, which superimposes two images with a certain transparency and combine both contraband and baggage simultaneously, results in seriously implausible representations because transparency in X-ray images is dependent on both material and thickness~\cite{bhowmik2019good}.
The reason behind the performance degradation lies in the difference between natural image domain and X-ray security image domain, it is difficult for detectors to learn high discriminative and occluded regions equally.

\textbf{\textit{Can we exploit the characteristics of X-ray security images to design a simple and effective data augmentation method?}}
An ideal data augmentation should improve the diversity of training samples, to ensure that the same instance is consistently recognized in various environments.
To achieve this, the simple and straightforward way is to simulate the rich background, which induces model to benefit from complex scenarios and reduces the risk of model overfitting.
Furthermore, given the scarcity and imbalance of discriminative information in X-ray security images, a thorough analysis of domain-specific knowledge and a careful method design are particularly important.
Therefore, leveraging the transmission characteristics and material-based pseudo-coloring of X-ray images, we design a simple and effective data augmentation approach to simulate the unique clutter and occlusion in X-ray security inspection scenarios.
Specifically, since the luggage often contains only a limited number of items, the background remains relatively simple, allowing models to overfit on high-contrast regions. We make full use of the physical properties of X-ray security images to simulate difficult samples in complex environments at the patch level.
Firstly, at texture level, we extract background patches and move them randomly, and finally perform local Mixup operations.
Patch-level texture information is helpful for models to focus on the causal feature of prohibited items, rather than only on highly discriminative feature.
Furthermore, at material level, we generate randomly colored patches of varying sizes, and then perform Mixup operations at random locations of the image.
This straightforward way simulates occlusions caused by diverse materials, effectively encouraging models to learn broader, more robust discriminative features.

As shown in Fig.~\ref{fig:IntroBGM}, we highlight the dilemma of natural data augmentation lacking X-ray security data characteristic driven, while pointing out that the potential of background information from X-ray security images has not been explored in this field. Notably, BGM is not merely an extension of Mixup, it is rooted in insights from X-ray data characteristics, offering a straightforward and effective solution with practical application, without additional annotations or significant training overhead.
\textit{\textbf{Concretely}},
SPM leverages local baggage texture to enhance scene complexity, a unique advantage in transmitted light environments.
CPM adds random color patches to fill missing material information, constructing more complex samples.
Therefore, both SPM and CPM are carefully designed for X-ray security image through in-depth analysis and observation, reducing the augmentation gap of common visual augmentations of natural image.
It is worth noting that unlike existing X-ray security image augmentation focused on foreground~\cite{mery2017logarithmic, chen2024augmentation}, our work \textbf{centers on background-based augmentation}, aimed at providing an implicit regularization technique to strengthen model generalization against the unique occlusions in the field.
In addition, BGM can \textbf{be combined with foreground-guided image augmentation} to provide richer data samples and to further improve detection performance.
In summary, our contributions are as follows:

\begin{itemize}
\item To the best of our knowledge, we are the first to \textbf{explore and highlight the significance of background information in X-ray prohibited items detection}, addressing a long-overlooked yet critical factor in X-ray security image analysis.
\item We propose a straightforward and effective data augmentation technique specifically for X-ray prohibited items detection, called \textbf{Background Mixup (BGM)}. By exploiting the characteristics of X-ray security image, we expand the augmentation space, complementary to that of classical augmentations.
\item Based on a comprehensive analysis of both X-ray transmission imagery and material-based pseudo coloring, we carefully simulate complex textural structures and abundant material information via patch‑level augmentation.
\item Extensive experiments on multiple models and datasets show that BGM consistently improves prohibited items detection performance, enhances detector generalization, and effectively mitigates the domain-specific challenges posed by X-ray imaging’s physical properties, \textbf{without requiring additional annotations or increasing training overhead}.
\end{itemize}

\begin{table}[t]
\caption{\textbf{Performance of natural image augmentation methods and BGM on PIDray~\cite{zhang2023pidray} with DINO-R50~\cite{zhang2022dino}.}~(RE denotes Random Erasing~\cite{zhong2020random}).}
\Huge
\centering
\resizebox{0.88\linewidth}{!}{%
\begin{tabular}{l|cc|cccc}
\toprule
\multirow{2}{*}{\textbf{Method}} & \multicolumn{2}{c|}{\textbf{Type}} & \multicolumn{4}{c}{\textbf{Detection mAP}} \\
\cmidrule(lr){2-3} \cmidrule(lr){4-7}
 & \textbf{Image} & \textbf{Inst.} & \textbf{Easy} & \textbf{Hard} & \textbf{Hidden} & \textbf{Overall} \\
\midrule
Baseline\cite{zhang2022dino}              & - & -            & 74.0 & 69.7 & 52.1 & 68.4 \\
\midrule
+RE\cite{zhong2020random}    & Single & \XSolidBrush  & 74.1 & 70.0 & 51.0 & 68.7 \\
+Mixup\cite{zhang2017mixup}              & Multi & \XSolidBrush & 75.4 & 68.2 & 45.1 & 66.7 \\
+CutMix\cite{yun2019cutmix}              & Multi & \XSolidBrush  &  73.3  &   68.5   &  52.1 & 67.7    \\

+Copy-Paste\cite{ghiasi2021simple}       & Multi & \CheckmarkBold   & 75.4 & 69.4 & 46.4 & 67.3 \\

\rowcolor[rgb]{0.988,0.914,0.914} \textbf{+BGM (Ours)} 
                                          & Single & \XSolidBrush   & \textbf{76.9} & \textbf{70.5} & \textbf{52.9} & \textbf{70.1} \\
\rowcolor[rgb]{0.988,0.914,0.914} \textbf{\textit{Improvement}} 
                                          & - & -             & \textbf{\textcolor[rgb]{0,0,0}{+2.9\%}} & \textbf{\textcolor[rgb]{0,0,0}{+0.8\%}} & \textbf{\textcolor[rgb]{0,0,0}{+0.8\%}} & \textbf{\textcolor[rgb]{0,0,0}{+1.7\%}} \\
\bottomrule
\end{tabular}}

\label{tab:augmentation_comparison}
\end{table}

\section{Related Work}
\label{sec:formatting}

This section concisely analyzes X-ray prohibited items detection related work and classical augmentations for natural image, aiming to demonstrate that the necessity of X-ray security image augmentation and the perspective of bridging the gap of the augmentation between natural image and X-ray security image.

\subsection{X-ray Prohibited Items Detection}

Current works on prohibited items detection basically focus on dataset construction ~\cite{tao2021towards,tao2022exploring,tao2022few,liu2023x,miao2019sixray,ma2024towards,wang2021towards,zhao2022detecting}, model improvement~\cite{zhu2024fdtnet, jia2024forknet, li2022pixdet, ma2024coarse}, and model-based data augmentation methods~\cite{liu2022data, mery2017logarithmic, duan2023rwsc, schwaninger2007adaptive}.
The construction of public datasets provides a convenient research way, and the model improvement work considers prior information or scene requirements for security inspection scenarios.
Model-based data augmentation methods~\cite{liu2022data, mery2017logarithmic, duan2023rwsc, schwaninger2007adaptive} tend to alleviate the challenges of labeling and data scarcity in model training by generating simulated images.
However, model-based augmentation methods demand significant computational and data resources, and their complex training processes, coupled with inherent data biases, pose challenges for flexible deployment.

Notably, TIP~\cite{bhowmik2019good}, as a carefully designed data augmentation technique designed for X-ray security images, differs from our approach in focus. 
TIP cuts out foreground regions of prohibited items and pastes them carefully into clear baggage images. The goal of the strategy is to increase the proportion of positive samples, to benefit detection models from additional loss, which is time-consuming and labor-intensive.

In contrast to introducing additional instances, our method aims at mining the physical properties of X-ray security images, 
operates on single baggage image, enhancing background diversity to help models handle severe occlusion and clutter at significant lower cost.
In addition, unlike foreground-based augmentation\cite{chen2024augmentation, webb2021operationalizing}, we explore augmentation strategies \textbf{from a background perspective without additional instances or training overhead}, compatible with foreground-aware augmentation strategies.

\subsection{Model-free Augmentation for Natural Images}

Model-free augmentation is applied to a single image or multiple images and improves model robustness and generalization, offering advantages such as parameter-free design, high efficiency, and simplicity~\cite{xu2023comprehensive}.
In natural image recognition, single image model-free augmentation includes rotation, flipping, Random Erasing~\cite{zhong2020random}, etc., while cross-image model-free augmentation includes Mixup~\cite{zhang2017mixup}, CutMix~\cite{yun2019cutmix}, Copy-Paste~\cite{ghiasi2021simple}, etc.
Random Erasing~\cite{zhong2020random} randomly covers a rectangular region, reducing the model’s dependency on specific local features, while no particularly outstanding performance is achieved in X-ray images of security scene.
Mixup blends two images and their labels in a certain proportion, enriching the training data distribution and providing smoother decision boundaries for model inference. 
In Mixup~\cite{zhang2017mixup}, the randomly selected X-ray image is scaled, flipped, and clipped according to the original image, which results in a large number of meaningless areas being embedded in the original image, seriously interfering with image quality and data distribution,  as illustrated in Table~\ref{tab:augmentation_comparison}, the performance is dropped compared with the baseline.
For CutMix~\cite{yun2019cutmix}, considering X-ray security image with less discriminative information, it is rough to simply introduce a local region from another image, with the limited performance. We give further verification and analysis in Sec.~\ref{sec:exp}.
Copy-Paste creates varied object combinations, enhancing the diversity and complexity of samples.
However, compared with TIP~\cite{bhowmik2019good}, which is a more refined foreground operation suitable specially for X-ray security image, Copy-Paste has limited improvement.

Consequently, it is challenging to directly transfer the success of natural image data augmentation techniques to the X-ray prohibited items detection task. 
There is a significant gap between natural and X-ray security images in terms of imaging technology and visual characteristics, with the latter requiring \textbf{a deep understanding of X-ray physical properties and more careful and refined design}. 
Current natural data augmentation techniques are seldom specifically designed for X-ray security images and often overlook their distinctive data characteristics. 
Within this field, existing data augmentation strategies typically involve introducing new instances or synthesizing samples to expand the dataset. 
Fundamentally, these methods aim to improve detection model performance by increasing training constraints. 
What's missing in existing work is a deeper exploration of X-ray image properties and the valuable background information they contain. 
Our paper fills this void by being the first to propose a simple and effective regularization technique based on background cues to alleviate detection difficulties in this domain. 
Crucially, our approach is rooted in the intrinsic data features, not merely in adding constraints, and it seamlessly integrates with foreground augmentation, thereby presenting a fresh paradigm for data-driven learning in prohibited items detection.

\begin{figure}[t]
   \centering   
   \includegraphics[width=0.95\linewidth]{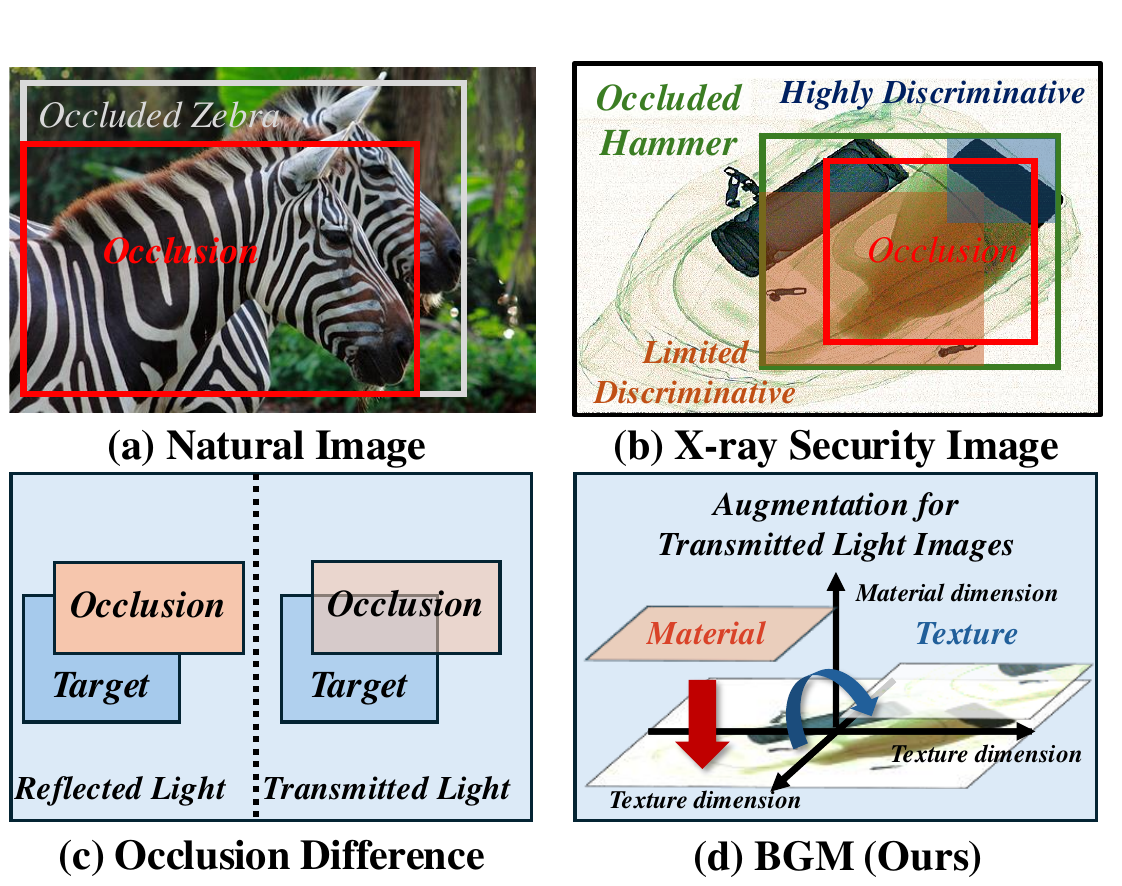}
   \caption{\textbf{Occlusion in two physical domains.} \textbf{(a)} Natural image is from reflected light imaging. \textbf{(b)} X-ray security image is from transmitted light imaging. The handle of the hammer is partially occluded by a container made of the similar material.
    Both highly and limited discriminative regions appear in one instance, resulting in the special inconsistency of semantic information. \textbf{(c)} occlusion difference between two physical domains. \textbf{(d)} our method stems from simulation of occlusion in the X-ray security images.}
   \label{fig:preliminary}   
\end{figure}
\section{Methodology}
In this section, we first comprehensively analyze the unique characteristics of X-ray security images, including transmission imaging and pseudo-color rendering techniques, and then give the problem definition in~\ref{sec:preliminary}. 
Secondly, we introduce a prohibited items detection framework that integrates our approach in~\ref{sec:framework}. 
Thirdly, we describe Background Mixup, a direct and effective data augmentation method to construct complex samples at both texture and material levels, including two strategies, SPM and CPM in~\ref{sec:spm} and~\ref{sec:cpm}.
Fourthly, we present the complete procedure of the method, ensuring clarity for the reader in~\ref{sec:procedure of bgm}. 
Finally, we discuss the differences between the proposed method and the previous augmentation methods in~\ref{sec:comparsion.pre.aug.}.

\subsection{Preliminary}
\label{sec:preliminary}

\subsubsection{X-ray Transmission Imagery}

Basically, X-ray security images are collected through X-ray imagery system, following by enhanced visualization through pseudo colorization~\cite{velayudhan2022recent}.
The final observed intensity in X-ray image depends on all the objects in the X-ray path. 
Hence, X-ray transmission images differ from reflection images, where a pixel in the image only belongs to a single item~\cite{velayudhan2022recent}.
Due to unique physical properties, X-ray images often lack texture details and have low contrast, with confusion and occlusion between objects.
As shown in Fig.~\ref{fig:preliminary}, the occlusion from reflected light imaging and X-ray exhibits extremely different appearances.

\subsubsection{Material-based Pseudo-coloring}

X-ray security inspection systems determine material composition using a look-up table calibrated by analyzing attenuation at specific energy levels. 
Pseudo-coloring images based on this information enhances the ability to distinguish baggage contents, where, for instance, organic materials typically appear orange, while high-density metals are shown in blue~\cite{velayudhan2022recent,akcay2022towards}, which enables better discrimination of multiple baggage contents(as shown in Fig.~\ref{fig:preliminary}~(b)).

\subsubsection{Problem Statement}
\begin{figure}[t]
  \centering  
   \includegraphics[width=0.95\linewidth]{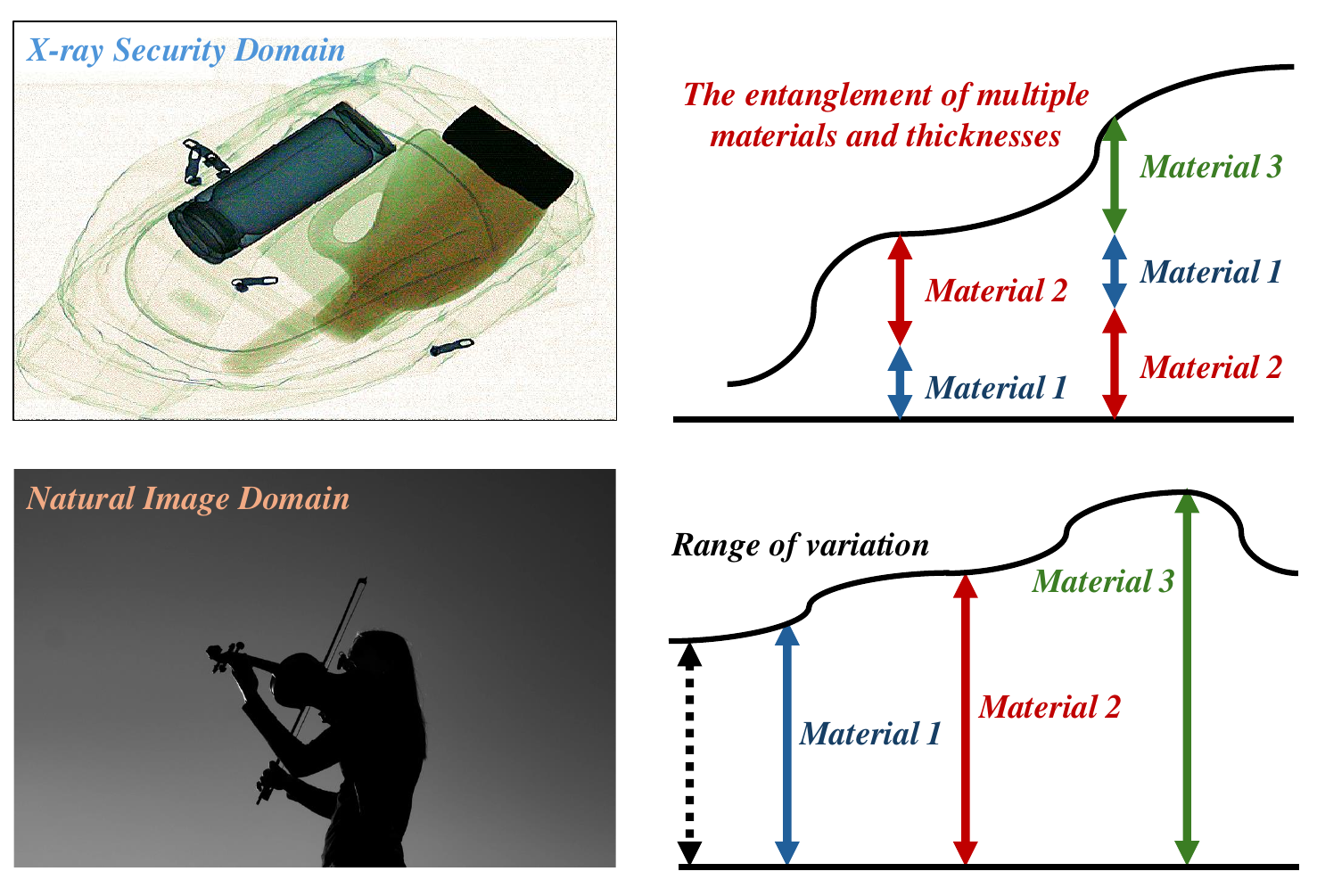}

 \caption{\textbf{Difference between two physical domains on the representation space of pixel value.} Different from the natural image domain, X-ray security images cover objects of various materials and thicknesses in a limited pixel representation range. The coupling of each other forms a unique occlusion in visualization, which is a domain-specific challenge.}
   \label{fig:suppl_problem_revisit}   
\end{figure}

We revisit the problem definition here to further illustrate domain-specific challenges.
In general, X-ray security images pose domain-specific challenges stemming from two physical principles: transmission imaging and pseudo-color rendering. 
These lead to unique occlusions caused by material properties and object thickness~\cite{velayudhan2022recent}. 
As shown in Fig.~\ref{fig:suppl_problem_revisit}, observations from publicly available datasets reveal that such occlusions arise from a compressed representation in the pixel space, where multiple materials and thicknesses coexist within a small region.
Notably, in natural images, even under low-light conditions, the pixel space, typically represents a single dominant object.
Meanwhile, considering that targets may be composed of different materials, targets containing both metal and organic materials will appear visually incoherent in X-ray security images
This distinction underscores the difficulty posed by X-ray security images, where domain-specific occlusions pose a substantial challenge.

\subsection{Framework}
\label{sec:framework}

As mentioned above, X-ray security images embody two distinct physical attributes: X-ray Transmission Imagery and Material-based Pseudo-coloring, which lead to a domain-specific phenomenon where foreground and background regions become entangled~\cite{gaus2024performance}. 
Therefore, models tend to overfit on highly discriminative regions, thereby neglecting low-discriminative areas.
Since the handle of the hammer and the baggage background are similar in material in Fig.~\ref{fig:preliminary}, the low contrast region limit model performance in Fig.~\ref{fig:exp_qualitativev2}.
From a data-driven perspective, encouraging models to broaden their focus and capture more generalized features presents a significant domain challenge.

\begin{algorithm}[t]
\caption{Background Mixup Procedure}
\label{alg:BackgroundMixup}

\LinesNotNumbered
\SetKwInOut{Input}{Input}
\SetKwInOut{Output}{Output}

\Input{
    Image $X$ with height $H$, width $W$, channels $C$; \\
    Ground truth bounding boxes $GT_{\text{box}}$; \\
    Number of SPM patches $n_{\text{patch}}$; \\
    Number of CPM patches $m_{\text{patch}}$; \\
    Transparency range $[\alpha_{\min}, \alpha_{\max}]$
}
\Output{Augmented image $X_{\text{final}}$}
\LinesNumbered

\textbf{Step 1: Random Component Selection} \\
Random choice from: SPM, CPM, or SPM + CPM\;

\textbf{Step 2: Patch Selection and Transformation} \\
\If{SPM is selected}{
    Select $n_{\text{patch}}$ background patches $P_i$ from $X$ excluding $GT_{\text{box}}$\;
    \For{each SPM patch $P_i$}{
        Apply random shifts $(\Delta x, \Delta y)$ and move $P_i$ to new position $(x_i + \Delta x, y_i + \Delta y)$ as $P_i'$\;
    }
}
\If{CPM is selected}{
    Select $m_{\text{patch}}$ random patches $P_k$ from $X$\;
    \For{each CPM patch $P_k$}{
        Assign random color $C_k = [c_{R_k}, c_{G_k}, c_{B_k}]$ with each $c_{i_k} \sim \mathcal{U}(0, 255)$\;
    }
}
\textbf{Step 3: Transparency Assignment and Mixup} \\
For each selected patch, sample a random transparency $\alpha \sim \mathcal{U}(\alpha_{\min}, \alpha_{\max})$\; 
\If{SPM is selected}{
    Apply Mixup: $X_{\text{mix}} = \alpha \cdot P' + (1 - \alpha) \cdot X_{(x', y')}$\;
}
\If{CPM is selected}{
    Apply Mixup: $X'(u, v) = (1 - \alpha) \cdot X(u, v) + \alpha \cdot C$\;
}

\textbf{Step 4: Combine Results} \\
Form the final augmented image $X_{\text{final}}$ by merging $X_{\text{mix}}$ and $X'$ (if both SPM and CPM are selected)\;

\Return $X_{\text{final}}$\;

\end{algorithm}

\begin{figure*}[t]
   \centering
   \includegraphics[width=1\linewidth]{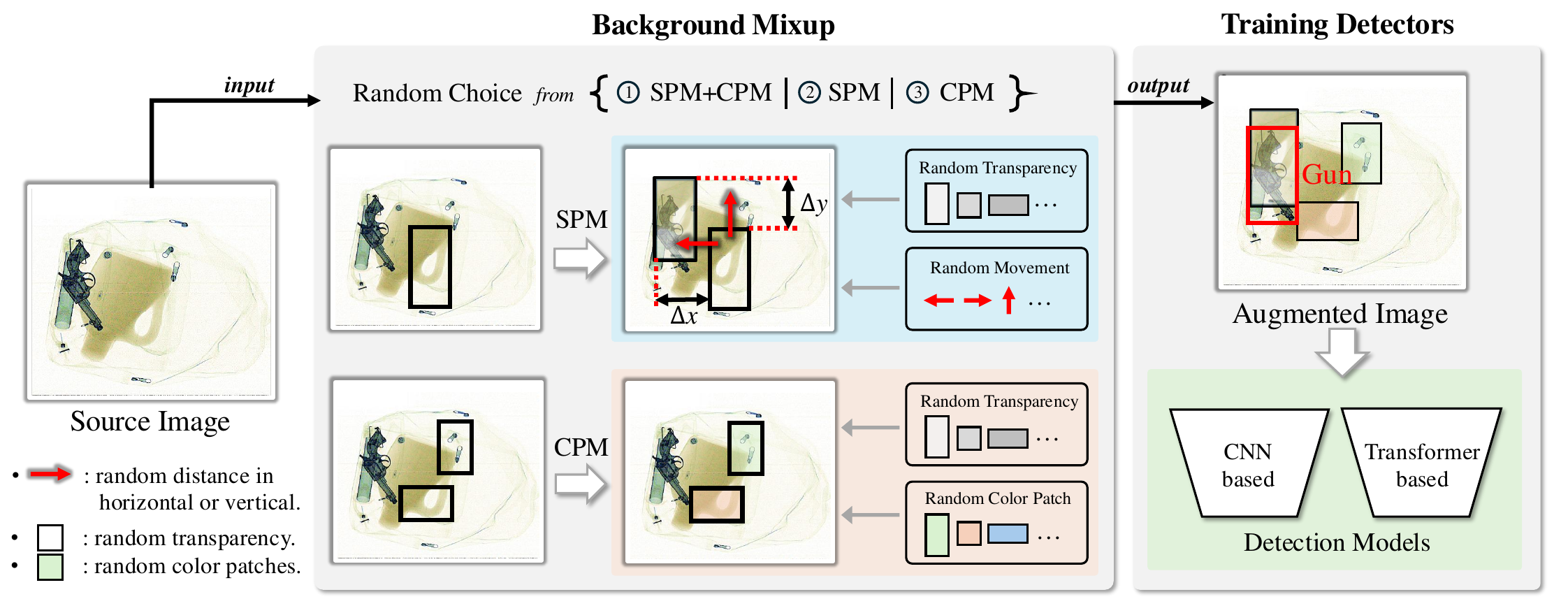}
   \caption{\textbf{Overview of the prohibited item detection framework integrated with BGM.} (1) When images and labels are prepared, we perform SPM for source input to carry out a flexible exploration of rich background information in a global image. Mixup operation is performed for randomly selected patches and images in the local area. (2) Meanwhile, for a more sufficient exploration, we perform CPM to provide material information in the X-ray image, which introduces several semi-transparent color patches to simulate variation of material. (3) Grounded in the careful and refined design driven by the physical domain, BGM is adapted to a variety of detection networks to help improve performance, including CNN-based detectors (e.g. Cascade R-CNN, ATSS) and Transformer-based detectors (e.g. DINO).}
   \label{fig:overview}
\end{figure*}

Following the previous observations, we start with the characteristics of the X-ray security data.
As shown in Fig.~\ref{fig:overview}, we explore X-ray security image augmentation approach by simulating complex background, i.e.
\textbf{Background Mixup~(BGM)}, including two kinds of data augmentation methods: \textbf{S}elf \textbf{P}atch \textbf{}{M}ixup (\textbf{SPM}) augmentation at contour level and \textbf{C}olor \textbf{P}atch \textbf{M}ixup (\textbf{CPM}) augmentation at material level. 
Firstly, because of the transmission property of X-ray images, coarse Mixup brings physically untrustworthy samples. Therefore, we design simple local operations to simulate real complex samples, so that the model can extract robust features in complex environments and resist occlusion and clutter in security inspection scenes.
SPM randomly selects patches from the baggage background (exclude the target foreground) and then randomly places them within the global scope to perform Mixup operation locally with a random transparency, thereby enriching the background information. 
Secondly, considering that SPM is flexible but may have difficulty incorporating different material information, CPM further introduces random color patches with random transparency to provide a more complex background. CPM randomly select several patches with random color, then perform Mixup operation locally within global scope of the X-ray image with a random transparency to provide additional information of material variation.
Both of two approaches introduce variations and simulated occlusions, enhancing the model’s robustness and improving generalization across different imaging scenarios. With our method integrated into the detection model framework, as illustrated in Fig.~\ref{fig:overview}, the two augmentation methods are randomly (independently or sequentially) applied to simulate rich background information. The random-choice strategy, which is evaluated (Table~\ref{tab:component_ablation}), aims to induce the model to enhance its attention to foreground objects by providing diverse and complex background contexts.

\subsection{Self Patch Mixup}
\label{sec:spm}
Let \( X \) represent an RGB image with height \( H \), width \( W \), and channels \( C \), acquired from an X-ray scanner and processed through proprietary denoising and pseudo-color rendering.
In the SPM process, given an image \( X \) and the ground truth bounding boxes \( GT_{\text{box}} \) representing the target foreground objects, the following steps are performed:

\noindent \textbf{Random Patch Selection.} Randomly select \( n_{\text{patch}} \) background patches from the image \( X \). Notably, the patches selected are in regions outside of the target foreground objects as defined by the ground truth boxes \( GT_{\text{box}} \). The purpose is to simulate X-ray baggage information close to reality, while rough movement of foreground may be harmful to the performance. The operation can be represented as:
\begin{equation}
\textit{Patches} = \{ P_i \mid i = 1, 2, \dots, n_{\text{patch}} \},
\end{equation}
where each \( P_i \) is a background patch, 
which is sampled from \( X \) except from \( GT_{\text{box}} \).

\noindent \textbf{Random Movement.} Each patch \( P_i \) with position \((x_i,y_i)\) is moved to a new position \( (x_i + \Delta x, y_i + \Delta y) \) within the global image area, where \( \Delta x \) and \( \Delta y \) are random horizontal and vertical shifts, respectively:
\begin{equation}
P_i' = P_i(x_i + \Delta x, y_i + \Delta y).   
\end{equation}
This movement allows the patches to cover various parts of the image \( X \) and introduces spatial diversity of contour. This operation is expected to simulate the real sample as much as possible by increasing the sample complexity.

\noindent \textbf{Random Transparency Assignment.} In order to simulate the transparency characteristic of  X-ray security image, assign each patch \( P_i' \) a transparency coefficient \( \alpha_i \), which is drawn from a predefined range \([ \alpha_{\min}, \alpha_{\max} ]\):
\begin{equation}
\alpha_i \sim \mathcal{U}(\alpha_{\min}, \alpha_{\max}).
\end{equation}

\noindent \textbf{Mixup Application.} A Mixup operation is applied to each patch \( P_i' \) and the corresponding region in \( X \) at its new position, using the random transparency coefficient \( \alpha_i \):

\begin{equation}
X_{\text{mix}} = \alpha_i \cdot P_i' + (1 - \alpha_i) \cdot X_{(x_i', y_i')},
\end{equation}
where \( X_{\text{mix}} \) is the resulting image and \( X_{(x_i', y_i')} \) represents the original region in \( X \) at the new position of \( P_i' \).

\noindent \textbf{Summary of Steps.} The final augmented image is constructed by applying the above operations iteratively on the patches. This method creates a complex but close to realistic background to make model robust and enhances the generalization capability of the detection model. Notably, our method is not a simple Mixup variant. We consider the source of occlusion in X-ray security images and use a direct method to carefully model complex texture structures.

\subsection{Color Patch Mixup}
\label{sec:cpm}

In the CPM process, given an image \( X \), where no annotation is required, the following steps are performed:

\noindent \textbf{Random Patch Selection.} Randomly select \( m_{\text{patch}} \) patches within the image \( X \). Compared with SPM, the operation doesn't need to exclude ground truth for random selection.
\begin{equation}
\textit{Patches} = \{ P_k \mid k = 1, \dots, m_{\text{patch}}  \},
\label{eq:patches}
\end{equation}
where each \( P_k \) is a randomly selected patch in the image.

\noindent \textbf{Random Color Assignment.} For each selected patch \( P_k \), assign a random color \( C_k \) to simulate varied material information, where each RGB channel value \( c_{R_k}, c_{G_k}, c_{B_k} \) is sampled independently from a uniform distribution:

\begin{equation}
\begin{aligned}
C_k &= [c_{R_k}, c_{G_k}, c_{B_k}], \\
c_{i_k} &\sim \mathcal{U}(0, 255), \quad i \in \{R, G, B\}.
\end{aligned}
\label{eq:color_assignment}
\end{equation}

\noindent \textbf{Random Transparency Assignment.} Assign a transparency coefficient \( \alpha_k \) to each patch with a random color, which is sampled from a predefined range \([ \alpha_{\min}, \alpha_{\max} ]\):
\begin{equation}
\alpha_k \sim \mathcal{U}(\alpha_{\min}, \alpha_{\max}).
\label{eq:transparency}
\end{equation}

\noindent \textbf{Mixup Application.} The color patches are applied to the image using alpha blending at their corresponding patch locations. For each pixel \( (u, v) \) within a selected patch \( P_k \), the blending operation can be defined as:
\begin{equation}
X'(u, v) = (1 - \alpha_k) \cdot X(u, v) + \alpha_k \cdot C_k,
\label{eq:mixup_application}
\end{equation}
where \( X(u, v) \) is the original pixel value, \( C_k \) is the randomly assigned color for patch \( P_k \), and \( \alpha_k \) is the transparency level sampled for this patch. Here, \( X'(u, v) \) represents the resulting augmented pixel value at location \( (u, v) \). 

\noindent \textbf{Summary of Steps.} Background patches are first selected, assigned random colors and transparency levels, then combined with the image using Mixup.  This operation simulates material variations in X-ray security images by overlaying random color patches with a semi-transparent effect. As shown in Fig.~\ref{fig:preliminary}, the two methods consider different dimensions of background information, SPM focuses on constructing rich texture structures, and CPM introduces additional material information from to provide further diversity of training samples.

\subsection{The Procedure of the Algorithm}
\label{sec:procedure of bgm}

To present our data augmentation method in a clearer and more standardized flow, the procedure for the proposed Background Mixup is outlined in Algorithm~\ref{alg:BackgroundMixup}. When an image and its corresponding labels are fed into the detection framework, we randomly select from three options: single component SPM, single component CPM, or the sequential combination SPM + CPM, applying augmentation at both the contour and material levels. Finally, standard model training is performed for X-ray prohibited items detection.

\subsection{Comparison against Previous Augmentation Methods.}
\label{sec:comparsion.pre.aug.}

Previous data augmentation strategies in X-ray security domain primarily focused on constructing foreground variations~\cite{mery2017logarithmic,bhowmik2019good,chen2024augmentation,webb2021operationalizing} to provide stronger supervision, i.e instance-level augmentation.
In contrast, we offer a new perspective by shifting the focus of data augmentation from the foreground to the background.
We argue that background-aware augmentation brings additional benefits for prohibited item detection in X-ray imagery. 
Combined with the regularization method, the detection model can deal with the special occlusion and clutter in the X-ray security inspection scene.
Our method can improve the detection performance to prevent missed detection and misclassification, especially for contraband with local low contrast.
Our experiments further show that BGM can be effectively combined with foreground-based methods to yield complementary gains.

Moreover, unlike traditional data augmentation techniques such as Random Erasing~\cite{zhong2020random}, which are originally designed for natural image scenarios, BGM is tailored for prohibited item detection in X-ray security imagery. It follows a data-driven learning paradigm by modeling background information through a patch-level augmentation mechanism, enabling the model to better exploit the complex and cluttered background context in security scenes.
In contrast to generic strategies like CutMix~\cite{yun2019cutmix} and MixUp~\cite{zhang2017mixup}, BGM specifically focuses on enhancing the representation of \textbf{background} information in X-ray security images~(also compatible with foreground-based augmentation methods). It further takes into account the domain-specific characteristics of the data when designing the augmentation process, thereby improving detection robustness and generalization.

\section{Experiments}
\label{sec:exp}
We firstly introduce the experimental setup including the dataset of the experiment, the model structure, the implementation details and the evaluation manner in~\ref{subsec:exp_setup}. 
Then we compare the detection performance with SOTA method and several detectors on PIDray~\cite{zhang2023pidray} in~\ref{subsec:exp_comp_sota}. 
Thirdly, we conduct component and hyperparameter ablation experiments in~\ref{subsec:exp_abla}. Fourthly, we further analyze the effectiveness of BGM on OPIXray~\cite{wei2020occluded} and CLCXray~\cite{zhao2022detecting} datasets, and conduct analysis on different Mixup Settings, color priors, instance segmentation, visualization and others in~\ref{subsec:exp_further}.
BGM not only offers substantial practical value, but also facilitates the model’s ability to capture generalized features in regions with highly imbalanced discriminative properties.

\begin{table*}[htbp]
\caption{\textbf{Comparisons on PIDray~\cite{zhang2023pidray}.} Detectors including the CNN-based architecture and Transformer-based architecture are used to evaluate the generalization of our approach. The high-performance baseline model, which integrates the propose simple enhancement approach, outperforms the best current public work on prohibited item detection. L1, L2, L3 denote different levels of detection difficulty, which mean easy, hard and hidden level, respectively. BA, PL, HA, PO, SC, WR, GU, BU, SP, HA, KN and LI denote Baton, Pliers, Hammer, Powerbank, Scissors, Wrench, Gun, Bullet, Sprayer, HandCuffs, Knife and Lighter in PIDray, respectively.}
\Large
\centering
\resizebox{1.0\textwidth}{!}{
\begin{tabular}{llll|cccc|ccccccccccccc}
\toprule
\multirow{2}{*}{\textbf{Type}} & \multirow{2}{*}{\textbf{Stage}} & \multirow{2}{*}{\textbf{Method}} & \multirow{2}{*}{\textbf{B-bone}} & \multicolumn{4}{c|}{\textbf{Detection mAP}} & \multicolumn{12}{c}{\textbf{AP Performance across Various Categories}} \\
\cmidrule(l){5-8} \cmidrule(l){9-20}
 &  &  &  & \textbf{L1} & \textbf{L2} & \textbf{L3} & \textbf{All} & \textbf{BA} & \textbf{PL} & \textbf{HA} & \textbf{PO} & \textbf{SC} & \textbf{WR} & \textbf{GU} & \textbf{BU} & \textbf{SP} & \textbf{HA} & \textbf{KN} & \textbf{LI} \\
\midrule

\multirow{9}{*}{CNN} & \multirow{2}{*}{One} & ATSS\cite{zhang2020bridging} & R101 & 71.7 & 65.8 & 47.9 & 65.2 & 71.9 & 81.6 & 76.3 & 74.0 & 71.9 & 84.1 & 25.9 & 60.8 & 59.1 & 84.4 & 32.9 & 59.3 \\  
 &  & \cellcolor[rgb]{0.988,0.914,0.914}\textbf{ATSS+BGM} & \cellcolor[rgb]{0.988,0.914,0.914}\textbf{R101} & \cellcolor[rgb]{0.988,0.914,0.914}\textbf{\textcolor[rgb]{0,0,0}{72.8}} & \cellcolor[rgb]{0.988,0.914,0.914}\textbf{\textcolor[rgb]{0,0,0}{66.4}} & \cellcolor[rgb]{0.988,0.914,0.914}\textbf{\textcolor[rgb]{0,0,0}{50.0}} & \cellcolor[rgb]{0.988,0.914,0.914}\textbf{\textcolor[rgb]{0,0,0}{66.4}} & \cellcolor[rgb]{0.988,0.914,0.914}\textbf{72.5} & \cellcolor[rgb]{0.988,0.914,0.914}\textbf{81.3} & \cellcolor[rgb]{0.988,0.914,0.914}\textbf{77.5} & \cellcolor[rgb]{0.988,0.914,0.914}\textbf{74.8} & \cellcolor[rgb]{0.988,0.914,0.914}\textbf{70.9} & \cellcolor[rgb]{0.988,0.914,0.914}\textbf{83.4} & \cellcolor[rgb]{0.988,0.914,0.914}\textbf{32.1} & \cellcolor[rgb]{0.988,0.914,0.914}\textbf{60.0} & \cellcolor[rgb]{0.988,0.914,0.914}\textbf{62.8} & \cellcolor[rgb]{0.988,0.914,0.914}\textbf{84.7} & \cellcolor[rgb]{0.988,0.914,0.914}\textbf{36.7} & \cellcolor[rgb]{0.988,0.914,0.914}\textbf{59.5} \\
\cmidrule(l){2-20}
 & \multirow{7}{*}{Two} & SDANet\cite{wang2021towards} & R101 & 72.2 & 63.7 & 48.0 & 64.4 & 71.0 & 81.5 & 78.8 & 71.9 & 69.2 & 86.1 & 33.4 & 57.4 & 60.2 & 84.8 & 30.4 & 52.6 \\   
 &  & Improved\cite{zhang2023pidray} & R101 & 74.5 & 64.8 & 53.0 & 66.6 & 72.9 & 83.2 & 78.3 & 73.2 & 70.2 & 86.1 & 39.3 & 58.4 & 61.5 & 85.6 & 35.4 & 54.8 \\ 
 & & C-RCNN\cite{cai2018cascade} & R101 & 74.7 & 68.2 & 51.8 & 68.0 & 73.5 & 83.1 & 79.8 & 75.0 & 73.7 & 88.4 & 33.1 & 63.4 & 61.2 & 86.4 & 42.1 & 56.7 \\
 &  & \cellcolor[rgb]{0.988,0.914,0.914}\textbf{C-RCNN+BGM} & \cellcolor[rgb]{0.988,0.914,0.914}\textbf{R101} & \cellcolor[rgb]{0.988,0.914,0.914}\textbf{\textcolor[rgb]{0,0,0}{75.3}} & \cellcolor[rgb]{0.988,0.914,0.914}\textbf{\textcolor[rgb]{0,0,0}{69.0}} & \cellcolor[rgb]{0.988,0.914,0.914}\textbf{\textcolor[rgb]{0,0,0}{52.8}} & \cellcolor[rgb]{0.988,0.914,0.914}\textbf{\textcolor[rgb]{0,0,0}{69.5}} & \cellcolor[rgb]{0.988,0.914,0.914}\textbf{75.2} & \cellcolor[rgb]{0.988,0.914,0.914}\textbf{84.0} & \cellcolor[rgb]{0.988,0.914,0.914}\textbf{81.4} & \cellcolor[rgb]{0.988,0.914,0.914}\textbf{75.3} & \cellcolor[rgb]{0.988,0.914,0.914}\textbf{74.6} & \cellcolor[rgb]{0.988,0.914,0.914}\textbf{88.9} & \cellcolor[rgb]{0.988,0.914,0.914}\textbf{38.4} & \cellcolor[rgb]{0.988,0.914,0.914}\textbf{64.6} & \cellcolor[rgb]{0.988,0.914,0.914}\textbf{61.9} & \cellcolor[rgb]{0.988,0.914,0.914}\textbf{87.1} & \cellcolor[rgb]{0.988,0.914,0.914}\textbf{44.8} & \cellcolor[rgb]{0.988,0.914,0.914}\textbf{57.8} \\
 
 \cmidrule(l){3-20}
 &  & FDTNet\cite{zhu2024fdtnet} & X101 & 77.2 & 69.6 & 57.9 & 68.2 & - & - & - & - & - & - & - & - & - & - & - & - \\
 &  & C-RCNN\cite{cai2018cascade} & X101 & 75.5 & 69.4 & 54.3 & 69.6 & 74.5 & 83.7 & 81.4 & 76.4 & 75.1 & 89.2 & 31.2 & 66.2 & 62.8 & 87.9 & 47.0 & 59.3 \\
 &  & \cellcolor[rgb]{0.988,0.914,0.914}\textbf{C-RCNN+BGM} & \cellcolor[rgb]{0.988,0.914,0.914}\textbf{X101} & \cellcolor[rgb]{0.988,0.914,0.914}\textbf{\textcolor[rgb]{0,0,0}{77.4}} & \cellcolor[rgb]{0.988,0.914,0.914}\textbf{\textcolor[rgb]{0,0,0}{70.3}} & \cellcolor[rgb]{0.988,0.914,0.914}\textbf{\textcolor[rgb]{0,0,0}{55.0}} & \cellcolor[rgb]{0.988,0.914,0.914}\textbf{\textcolor[rgb]{0,0,0}{70.6}} & \cellcolor[rgb]{0.988,0.914,0.914}\textbf{75.6} & \cellcolor[rgb]{0.988,0.914,0.914}\textbf{84.1} & \cellcolor[rgb]{0.988,0.914,0.914}\textbf{81.2} & \cellcolor[rgb]{0.988,0.914,0.914}\textbf{77.0} & \cellcolor[rgb]{0.988,0.914,0.914}\textbf{75.4} & \cellcolor[rgb]{0.988,0.914,0.914}\textbf{89.0} & \cellcolor[rgb]{0.988,0.914,0.914}\textbf{39.6} & \cellcolor[rgb]{0.988,0.914,0.914}\textbf{66.3} & \cellcolor[rgb]{0.988,0.914,0.914}\textbf{63.4} & \cellcolor[rgb]{0.988,0.914,0.914}\textbf{87.7} & \cellcolor[rgb]{0.988,0.914,0.914}\textbf{49.0} & \cellcolor[rgb]{0.988,0.914,0.914}\textbf{58.7} \\
\midrule

\multirow{4}{*}{Transformer} & & DINO\cite{zhang2022dino} & R50 & 74.0 & 69.7 & 52.1 & 68.4 & 76.2 & 86.1 & 83.9 & 74.8 & 72.1 & 90.6 & 29.6 & 62.2 & 56.2 & 89.6 & 38.7 & 61.0 \\
 &  & \cellcolor[rgb]{0.988,0.914,0.914}\textbf{DINO+BGM} & \cellcolor[rgb]{0.988,0.914,0.914}\textbf{R50} & \cellcolor[rgb]{0.988,0.914,0.914}\textbf{\textcolor[rgb]{0,0,0}{76.9}} & \cellcolor[rgb]{0.988,0.914,0.914}\textbf{\textcolor[rgb]{0,0,0}{70.5}} & \cellcolor[rgb]{0.988,0.914,0.914}\textbf{\textcolor[rgb]{0,0,0}{52.9}} & \cellcolor[rgb]{0.988,0.914,0.914}\textbf{\textcolor[rgb]{0,0,0}{70.1}} & \cellcolor[rgb]{0.988,0.914,0.914}\textbf{76.9} & \cellcolor[rgb]{0.988,0.914,0.914}\textbf{86.6} & \cellcolor[rgb]{0.988,0.914,0.914}\textbf{85.3} & \cellcolor[rgb]{0.988,0.914,0.914}\textbf{74.5} & \cellcolor[rgb]{0.988,0.914,0.914}\textbf{72.7} & \cellcolor[rgb]{0.988,0.914,0.914}\textbf{92.2} & \cellcolor[rgb]{0.988,0.914,0.914}\textbf{32.6} & \cellcolor[rgb]{0.988,0.914,0.914}\textbf{64.2} & \cellcolor[rgb]{0.988,0.914,0.914}\textbf{62.6} & \cellcolor[rgb]{0.988,0.914,0.914}\textbf{90.0} & \cellcolor[rgb]{0.988,0.914,0.914}\textbf{43.4} & \cellcolor[rgb]{0.988,0.914,0.914}\textbf{60.2} \\ 
 &  & DINO\cite{zhang2022dino} & Swin & 82.8 & 76.1 & 59.2 & 76.1 & 81.3 & 89.6 & 86.3 & 81.7 & 79.2 & 92.4 & 46.6 & 68.3 & 74.9 & 91.0 & 57.7 & 64.3 \\
 &  & \cellcolor[rgb]{0.988,0.914,0.914}\textbf{DINO+BGM} & \cellcolor[rgb]{0.988,0.914,0.914}\textbf{Swin} & \cellcolor[rgb]{0.988,0.914,0.914}\textbf{\textcolor[rgb]{0,0,0}{84.2}} & \cellcolor[rgb]{0.988,0.914,0.914}\textbf{\textcolor[rgb]{0,0,0}{76.6}} & \cellcolor[rgb]{0.988,0.914,0.914}\textbf{\textcolor[rgb]{0,0,0}{60.6}} & \cellcolor[rgb]{0.988,0.914,0.914}\textbf{\textcolor[rgb]{0,0,0}{77.4}} & \cellcolor[rgb]{0.988,0.914,0.914}\textbf{82.4} & \cellcolor[rgb]{0.988,0.914,0.914}\textbf{90.1} & \cellcolor[rgb]{0.988,0.914,0.914}\textbf{87.4} & \cellcolor[rgb]{0.988,0.914,0.914}\textbf{80.8} & \cellcolor[rgb]{0.988,0.914,0.914}\textbf{80.6} & \cellcolor[rgb]{0.988,0.914,0.914}\textbf{93.2} & \cellcolor[rgb]{0.988,0.914,0.914}\textbf{50.7} & \cellcolor[rgb]{0.988,0.914,0.914}\textbf{69.3} & \cellcolor[rgb]{0.988,0.914,0.914}\textbf{77.0} & \cellcolor[rgb]{0.988,0.914,0.914}\textbf{91.7} & \cellcolor[rgb]{0.988,0.914,0.914}\textbf{61.2} & \cellcolor[rgb]{0.988,0.914,0.914}\textbf{64.7} \\
 
\bottomrule
\end{tabular}}
\label{tab:pidray_performance}
\end{table*}

\subsection{Experimental Setup}
\label{subsec:exp_setup}

\noindent \textbf{Datasets.} To verify the generalization of the propose method, we train detection models on several influential benchmark datasets, namely PIDray\cite{wang2021towards,zhang2023pidray}, CLCXray\cite{zhao2022detecting} and  OPIXray\cite{wei2020occluded}. Notably, the evaluation datasets come from different institutions, different X-ray imaging acquisition devices, and different acquisition scenarios, which is different from the limited dataset evaluation of previous work, thereby it is a challenging evaluation to verify the generalization of our method. 
\textbf{PIDray}~\cite{wang2021towards}, \textbf{Prohibited Item Detection dataset}, presents further challenges to research regarding intentionally concealed threats. The dataset comprising 47,677 baggage scans with 12 categories of prohibited items holds the most extensive collection of baggage threat X-ray scans. The testing subset encompassing 40\% of the images is further divided into three subgroups: easy (9,482 single threat scans), hard (3,733 multiple threat scans), and hidden (5,005 deliberately concealed threat scans). 
\textbf{OPIXray}~\cite{wei2020occluded}, \textbf{Occluded Prohibited Items X-ray benchmark} released in 2020, is generated synthetically using software with standard baggage scans as background. It is comprised of 8885 baggage scans with different variety of cutters- folding knives (1993 images), straight knives (1044 images), utility knives (1978 images), multi-tool knives (1978 images), and scissors (1863 images). Around 30 training scans and five testing scans contain multiple threat objects.
\textbf{CLCXray}~\cite{zhao2022detecting}, \textbf{Cutters and liquid containers X-ray dataset}, contains 9,565 X-ray images, in which 4,543 X-ray images (real data) are obtained from the real subway scene and 5,022 X-ray images (simulated data) are scanned from manually designed baggage. There are 12 categories in the CLCXray dataset, including 5 types of cutters and 7 types of liquid containers. Five kinds of cutters include blade, dagger, knife, scissors, Swiss Army knife. Seven kinds of liquid containers include cans, carton drinks, glass bottle, plastic bottle, vacuum cup, spray cans, tin.

\noindent \textbf{Architectures and implementation details.}
Our approach can be integrated into many mainstream detection frameworks and datasets to evaluate its generalization across different backbone networks, detectors, multiple devices, multiple scenarios, and multiple contraband types of datasets. 
Specifically, we use MMDetection~\cite{mmdetection} framework to perform different detectors' implementation and to construct our method.
CNN-based detectors and Transformer-based detectors are trained to verify the generalization.
To ensure resolution-agnostic operation, the patch size is controlled by $R_{area}$, generates random scaling factors between 0 and 1.
The scaling factors multiply the width and height of the image to obtain the width and height, position, and moving distance of the patch.
To facilitate localization of the background region, we use a simple threshold saliency detection to avoid white region patches.
For fair comparisons, we train different detectors for 12 epochs and follow the same parameter configuration according to the official version provided by MMDetection~\cite{mmdetection}.
All experiments are conducted on 4 Nvidia RTX 3090 GPUs.

\noindent \textbf{Evaluation.}
The model performance is evaluated by MS-COCO metrics~\cite{lin2014microsoft}, using mean Average Precision (mAP) at a series of IoU thresholds and Average Precision (AP) for each category on all datasets.
The mean average precision (mAP) is widely used evaluation metric for the prohibited item detection. It quantifies a model’s precision-recall trade-off across varying confidence thresholds, offering a comprehensive assessment of its detection performance.
We evaluate the model performance using mAP metric for object detection and the Intersection over Union~(IoU) threshold is set from 0.5 to 0.95 with a step size of 0.05, and the results are averaged. 
Furthermore, we select the best-performing model to calculate the AP for each category to observe performance improvements across classes.
Additionally, we evaluate the difference in our method the performance of BGM on different occlusion levels of test data on PIDray~\cite{wang2021towards,zhang2023pidray}.

\begin{figure}[t]
  \centering
   \includegraphics[width=1.0\linewidth]{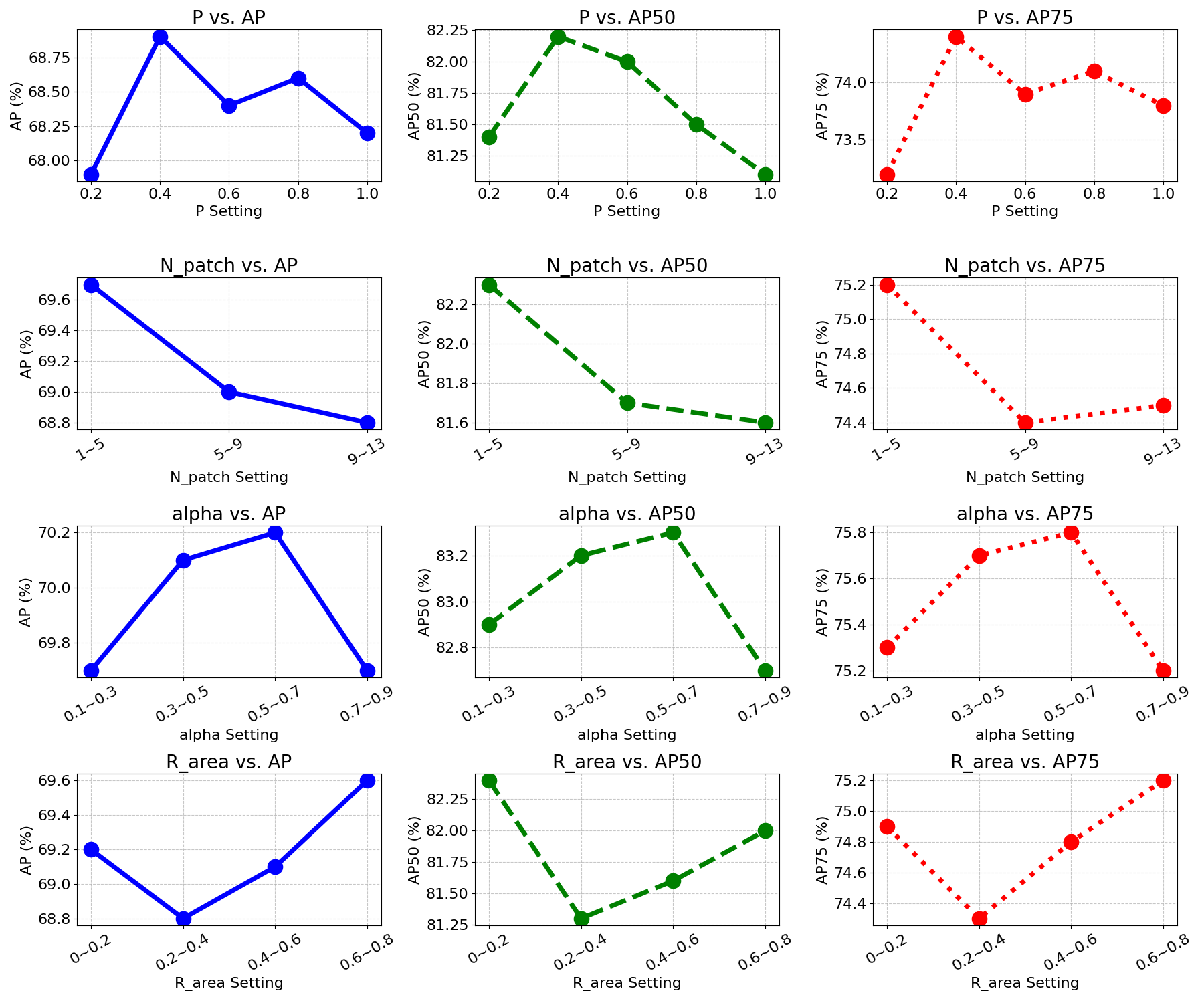}
   \caption{\textbf{Ablation study on hyper-parameters of SPM.}~$P$ denotes the probability of applying the strategy, $N_{patch}$ represents the range of patch numbers, $\alpha$ indicates the transparency range of patches, and $R_{area}$ denotes the range of patch area and location ratios.}
   \label{fig:SPM_Hyper_Setting}   
\end{figure}

\begin{table}[htbp]
\caption{\textbf{Ablation study of the proposed method on component analysis.} Single-Component means SPM or CPM, integrated with the detector. Random-Choice means random choice from SPM, CPM and Series Combination during training pipeline.}
\large
\centering
\resizebox{0.46\textwidth}{!}{
\begin{tabular}{l|cc|p{1.0cm}cc}
\toprule
\multirow{2}{*}{\textbf{Setting}} & \multicolumn{2}{c|}{\textbf{Module}} & \multicolumn{3}{c}{\textbf{Detection Performance}} \\
\cmidrule(lr){2-3} \cmidrule(lr){4-6}
 & \textbf{SPM} & \textbf{CPM} & \textbf{AP} & \textbf{AP\textsubscript{50}} & \textbf{AP\textsubscript{75}} \\
\midrule
Baseline & \XSolidBrush & \XSolidBrush & 68.4 & 81.7 & 73.9 \\
Single Component & \CheckmarkBold & \XSolidBrush & 69.8 & 82.6 & 75.5 \\
Single Component & \XSolidBrush & \CheckmarkBold & 69.8 & 82.4 & 75.4 \\
Series & \CheckmarkBold & \CheckmarkBold & 69.9  & 82.5 & 75.7 \\
\rowcolor[rgb]{0.988,0.914,0.914} Random Combination & \CheckmarkBold & \CheckmarkBold & \textbf{70.1} & \textbf{83.0} & \textbf{75.7} \\
\bottomrule
\end{tabular}}
\label{tab:component_ablation}
\end{table}

\subsection{Comparison with State of the art methods}
\label{subsec:exp_comp_sota}

We compare our method, BGM, with recent state of the art~(SOTA) methods with the same backbone and detector on PIDray~\cite{zhang2023pidray}. 
As shown in Table~\ref{tab:pidray_performance}, our method obtains gains of 1.7 and 1.3 on overall mAP over the baseline DINO~\cite{zhang2022dino}, with backbone of ResNet-50~\cite{he2016deep} and Swin~\cite{liu2021swin}, respectively. 
Furthermore, BGM, based on strong baselines,such as ATSS~\cite{zhang2020bridging}, DINO~\cite{zhang2022dino} with ResNet and Swin, Cascade-RCNN~\cite{cai2018cascade} with ResNet and ResNext, can consistently achieve SOTA performance, providing better performance on almost all metrics and demonstrating good effectiveness and generalization. 
Notably, the works for prohibited items detection are basically focusing on model's structure, which have parameters and computational time costs. 
Differently, we do not introduce learnable units or additional model structure.
Moreover, our method has achieved performance improvement in PIDray~\cite{zhang2023pidray} with multiple occlusion levels, showing a good ability to handle special occlusion in the X-ray security domain.

\subsection{Ablation Study}
\label{subsec:exp_abla}
We conduct ablation studies to study the effectiveness of each component in our method. Furthermore, we conduct hyper-parameter ablation study in two strategies, SPM and CPM.
Ablation studies are conducted on PIDray dataset~\cite{wang2021towards, zhang2023pidray}, with DINO-R50~\cite{zhang2022dino} detection model.

\subsubsection{Component-wise Ablation}

As shown in Table~\ref{tab:component_ablation}, we utilize baseline, single-component, series and random-choice setups to comprehensively evaluate the impact of our method. The result shows that the random-choice strategy yields superior performance, where one augmentation is randomly selected from two single components~(SPM and CPM) and their sequential combination. The results show that the random selection strategy can provide more rich enhancement transformations to help the model improve the detection accuracy from complex backgrounds.

\subsubsection{Hyper-parameter Ablation}
\begin{figure}[t]
  \centering
   \includegraphics[width=1.0\linewidth]{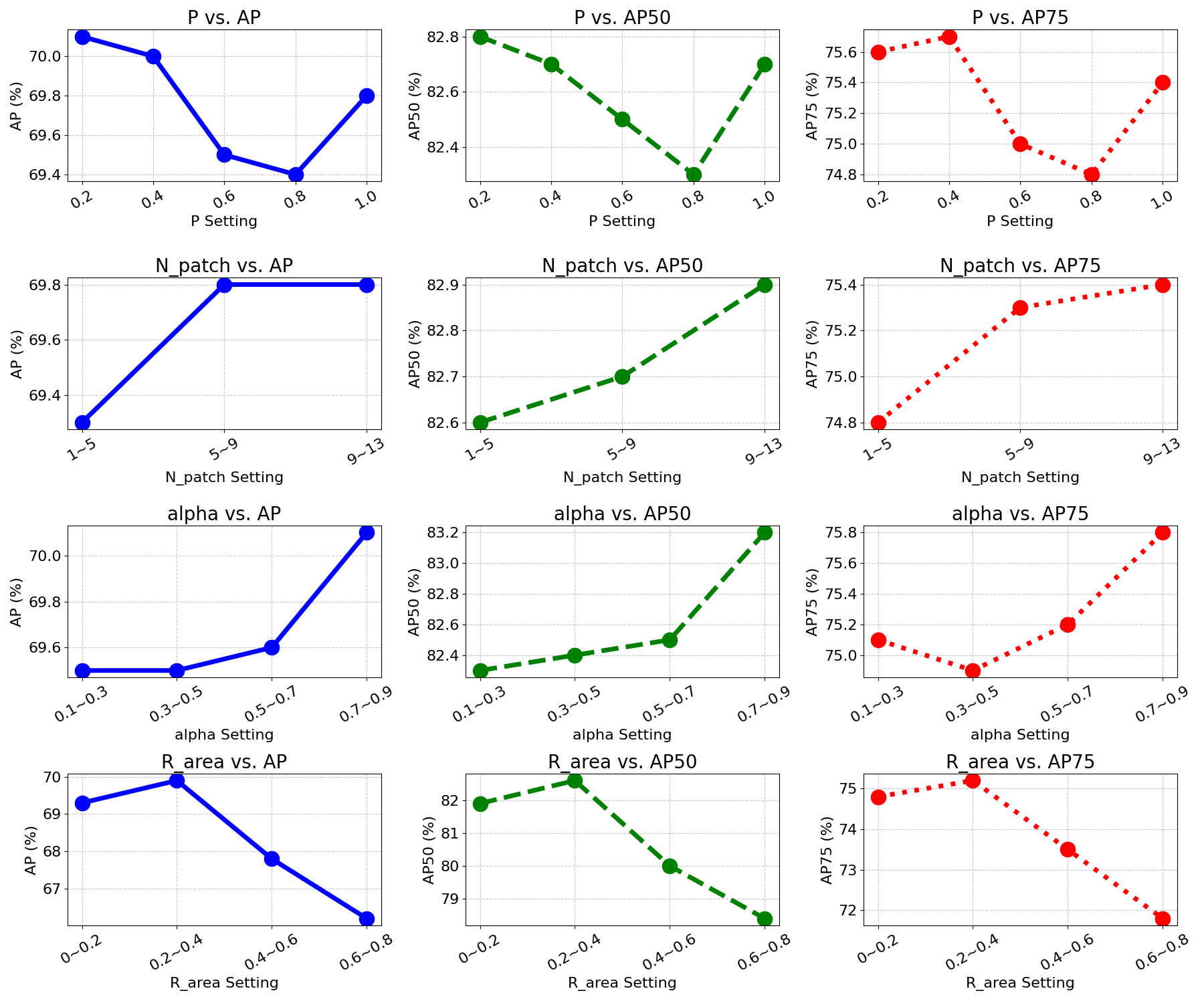}
   \caption{\textbf{Ablation study on hyper-parameters of CPM.}~$P$ denotes the probability of applying the strategy, $N_{patch}$ represents the range of patch numbers, $\alpha$ indicates the transparency range of patches, and $R_{area}$ denotes the range of patch area ratios.}
   \label{fig:CPM_Hyper_Setting}   
\end{figure}

We investigate the influence of various hyper-parameters on detection performance, including the probability of application, the range of patch numbers, area ratio, and transparency. 
To isolate each parameter’s effect, all other parameters are held constant during analysis.
The results of SPM and CPM are presented in Fig.~\ref{fig:SPM_Hyper_Setting} and Fig.~\ref{fig:CPM_Hyper_Setting}, respectively.
Our experiments reveal that SPM and CPM benefit from similar execution probability, benefiting from different number of patches, transparency ranges and scale parameters.
This indicates that proper addition of texture information and material information can help to improve the richness of training samples, and considering that texture information already covers complex structures, the optimal transparency of SPM is less than that of CPM.
Moreover, CPM performs better in a small scale range, which comes from the fact that CPM simulates objects of different materials, while SPM can provide more texture results in a large scale range.

\begin{figure*}[t]  
  \centering
  \includegraphics[width=\linewidth]{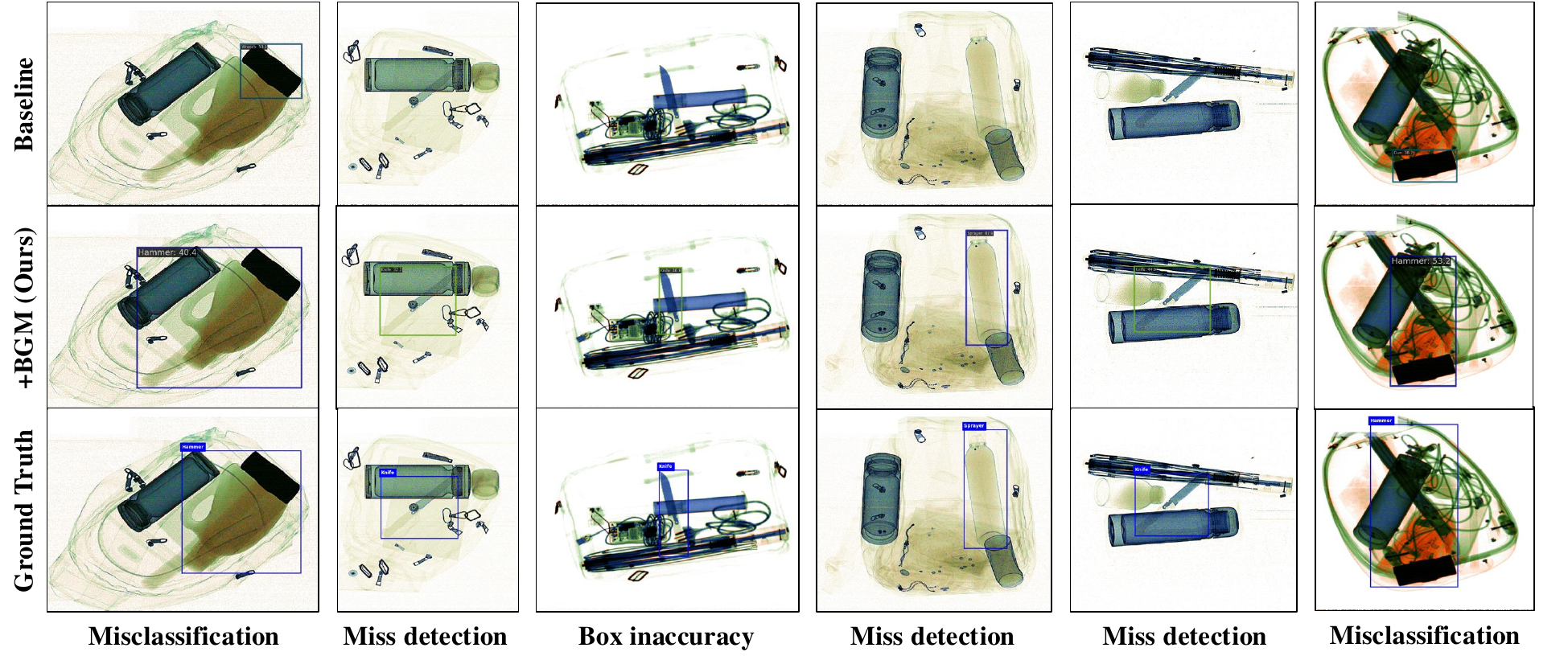}  
  \caption{\textbf{Qualitative examples} of BGM on PIDray~\cite{zhang2023pidray} dataset using DINO-R50~\cite{zhang2022dino}. BGM mitigates critical error cases including missed detections, misclassifications, and localization inaccuracies. BGM demonstrates strong detection performance on objects with low-contrast regions.}
  \label{fig:exp_qualitative_full}  
\end{figure*}

\begin{table}[htbp]
\caption{Performance between baseline~(DINO-R50~\cite{zhang2022dino}) and BGM on PIDray~\cite{wang2021towards,zhang2023pidray}, OPIXray~\cite{wei2020occluded} and CLCXray~\cite{zhao2022detecting}.}
\Large
\centering
\resizebox{0.46\textwidth}{!}{
\begin{tabular}{l|l|ccc}
\toprule
\multirow{2}{*}{\textbf{X-ray Dataset}} & \multirow{2}{*}{\textbf{Model Setting}} & \multicolumn{3}{c}{\textbf{Detection Performance}} \\
\cmidrule(l){3-5}
 & & \textbf{AP} & \textbf{AP$_{50}$} & \textbf{AP$_{75}$} \\
\midrule
 \multirow{3}{*}{PIDray\cite{zhang2023pidray,wang2021towards}} & Baseline & 68.4 & 81.7 & 73.9 \\
 & \cellcolor[rgb]{0.988,0.914,0.914}\textbf{+BGM} & \cellcolor[rgb]{0.988,0.914,0.914}\textbf{70.1} & \cellcolor[rgb]{0.988,0.914,0.914}\textbf{83.0} & \cellcolor[rgb]{0.988,0.914,0.914}\textbf{75.7} \\
 & \cellcolor[rgb]{0.988,0.914,0.914}\textit{Improvement} & \cellcolor[rgb]{0.988,0.914,0.914}\textcolor[rgb]{0,0,0}{+1.7\%} & \cellcolor[rgb]{0.988,0.914,0.914}\textcolor[rgb]{0,0,0}{+1.3\%} & \cellcolor[rgb]{0.988,0.914,0.914}\textcolor[rgb]{0,0,0}{+1.6\%} \\
\midrule
\multirow{3}{*}{OPIXray\cite{wei2020occluded}} & Baseline & 39.5 & 90.2 & 26.0 \\
 & \cellcolor[rgb]{0.988,0.914,0.914}\textbf{+BGM} & \cellcolor[rgb]{0.988,0.914,0.914}\textbf{40.4} & \cellcolor[rgb]{0.988,0.914,0.914}\textbf{91.0} & \cellcolor[rgb]{0.988,0.914,0.914}\textbf{27.7} \\
 & \cellcolor[rgb]{0.988,0.914,0.914}\textit{Improvement} & \cellcolor[rgb]{0.988,0.914,0.914}\textcolor[rgb]{0,0,0}{+0.9\%} & \cellcolor[rgb]{0.988,0.914,0.914}\textcolor[rgb]{0,0,0}{+0.8\%} & \cellcolor[rgb]{0.988,0.914,0.914}\textcolor[rgb]{0,0,0}{+1.7\%} \\
\midrule
\multirow{3}{*}{CLCXray\cite{zhao2022detecting}} & Baseline & 59.5 & 70.7 & 68.1 \\
 & \cellcolor[rgb]{0.988,0.914,0.914}\textbf{+BGM} & \cellcolor[rgb]{0.988,0.914,0.914}\textbf{61.4} & \cellcolor[rgb]{0.988,0.914,0.914}\textbf{72.6} & \cellcolor[rgb]{0.988,0.914,0.914}\textbf{68.6} \\
 & \cellcolor[rgb]{0.988,0.914,0.914}\textit{Improvement} & \cellcolor[rgb]{0.988,0.914,0.914}\textcolor[rgb]{0,0,0}{+1.9\%} & \cellcolor[rgb]{0.988,0.914,0.914}\textcolor[rgb]{0,0,0}{+1.9\%} & \cellcolor[rgb]{0.988,0.914,0.914}\textcolor[rgb]{0,0,0}{+0.5\%} \\
\bottomrule
\end{tabular}}
\label{tab:dataset_comparison}
\end{table}

\subsection{Further Analysis}
\label{subsec:exp_further}

\subsubsection{Extension to OPIXray and CLCXray}


To assess the generalization capability of BGM, we conduct evaluations across multiple X-ray security datasets, including PIDray~\cite{wang2021towards, zhang2023pidray}, OPIXray~\cite{wei2020occluded} and CLCXray~\cite{zhao2022detecting}. Detection performance is compared under two settings: with and without the integration of our proposed augmentation, BGM. The results, summarized in Table~\ref{tab:dataset_comparison}, demonstrate that BGM consistently improves detection performance over the baseline~\cite{zhang2022dino} across diverse datasets. 
It's worth noting that generalizing across scenario device datasets is challenging.
These datasets cover a wide range of scenarios, including station checkpoints and airports, various types of contraband, different X-ray imaging equipment, and distinct pseudo-color rendering techniques.

\begin{table}[htbp]
\caption{Performance of DINO-R50~\cite{zhang2022dino} on additional augmentation settings on PIDray~\cite{zhang2023pidray}. Our background-oriented augmentation method is compatible with existing foreground-based technique.}
\centering
\resizebox{0.46\textwidth}{!}{
\begin{tabular}{l|ccc}
\toprule
\multirow{2}{*}{\textbf{Setting}} & \multicolumn{3}{c}{\textbf{Detection Performance}} \\
\cmidrule(l){2-4}
 & \textbf{AP} & \textbf{AP$_{50}$} & \textbf{AP$_{75}$} \\
\midrule
Baseline & 68.4 & 81.7 & 73.9 \\
+Aug. on Multiple Images & 68.4 & 81.8 & 73.8 \\
+Aug. on Foreground Mixup & 59.5 & 74.0 & 64.1 \\
\rowcolor[rgb]{0.988,0.914,0.914} \textbf{+BGM} & \textbf{70.1} & \textbf{83.0} & \textbf{75.7}\\
\rowcolor[rgb]{0.988,0.914,0.914} \textit{\textbf{Improvement}} & \textbf{+1.6\%} & \textbf{+1.3\%} & \textbf{+1.8\%}\\
+TIP (Foreground Aug.)\cite{bhowmik2019good} & 70.6 & 83.1 & 76.1 \\
\rowcolor[rgb]{0.988,0.914,0.914} \textbf{+TIP \& BGM} & \textbf{71.7} & \textbf{84.1} & \textbf{77.3}\\
\rowcolor[rgb]{0.988,0.914,0.914} \textit{\textbf{Improvement}} & \textbf{+1.1\%} & \textbf{+1.0\%} & \textbf{+1.2\%}\\
\bottomrule
\end{tabular}}

\label{tab:detection_map_comparison}
\end{table}

\subsubsection{Extension to Multiple Images and Foreground}

We further extend our method by selecting background patches from one image and performing patch-level Mixup in another. As shown in Table~\ref{tab:detection_map_comparison}, the observed performance decline may be attributed to the significant variation in luggage content across images, which highlights the necessity of our carefully designed, context-aware augmentation strategy.

In addition, we investigate whether increasing the richness of foreground content could benefit detection performance. To this end, we locally move foreground regions within the image and apply Mixup operations. The resulting performance degradation, presented in Table~\ref{tab:detection_map_comparison}, is likely caused by the coarse manipulation of ground truth annotations, which disrupts the underlying distribution of training samples, which is consistent with observations in previous studies~\cite{webb2021operationalizing}.

\begin{table}[h]
\caption{\textbf{Different selection strategies.} Random color space has a wider selection of colors and has better generalization across devices with comparable performance.}
\normalsize
\centering
\resizebox{0.46\textwidth}{!}{
\begin{tabular}{l|l|p{0.8cm}cc}
\toprule
\multirow{2}{*}{\textbf{X-ray Dataset}} & \multirow{2}{*}{\textbf{Model Setting}} & \multicolumn{3}{c}{\textbf{Detection Performance}} \\
\cmidrule(l){3-5}
 & & \textbf{AP} & \textbf{AP$_{50}$} & \textbf{AP$_{75}$} \\
\midrule
\multirow{2}{*}{PIDray~\cite{zhang2023pidray,wang2021towards}}
    & Color Prior    & 69.8 & 82.7 & 75.4 \\
    & \cellcolor[rgb]{0.988,0.914,0.914}\textbf{+BGM}     & \cellcolor[rgb]{0.988,0.914,0.914}\textbf{70.1} & \cellcolor[rgb]{0.988,0.914,0.914}\textbf{83.0} & \cellcolor[rgb]{0.988,0.914,0.914}\textbf{75.7} \\
\midrule
\multirow{2}{*}{OPIXray~\cite{wei2020occluded}}    
    & Color Prior    & 40.0 & 91.4 & 25.4 \\
    & \cellcolor[rgb]{0.988,0.914,0.914}\textbf{+BGM}     & \cellcolor[rgb]{0.988,0.914,0.914}\textbf{40.4} & \cellcolor[rgb]{0.988,0.914,0.914}\textbf{91.0} & \cellcolor[rgb]{0.988,0.914,0.914}\textbf{27.7} \\
\midrule
\multirow{2}{*}{CLCXray~\cite{zhao2022detecting}}    
    & Color Prior    & 59.5 & 71.2 & 67.9 \\
    & \cellcolor[rgb]{0.988,0.914,0.914}\textbf{+BGM}     & \cellcolor[rgb]{0.988,0.914,0.914}\textbf{61.4} & \cellcolor[rgb]{0.988,0.914,0.914}\textbf{72.6} & \cellcolor[rgb]{0.988,0.914,0.914}\textbf{68.6} \\
\bottomrule
\end{tabular}}

\label{tab:main_color_statistics}
\end{table}

\begin{figure}[htbp]    
  \centering
  \includegraphics[width=0.49\textwidth]{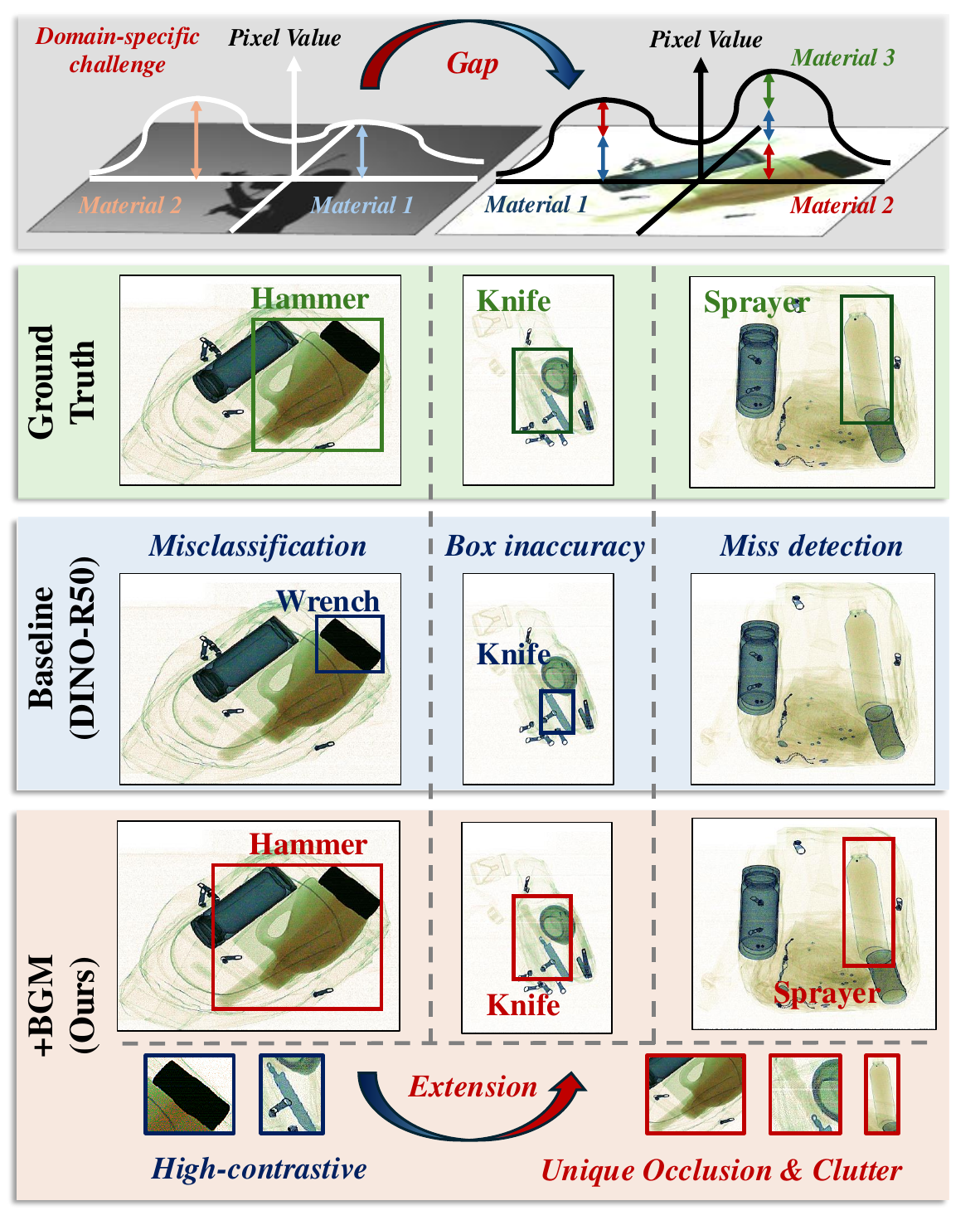}
  \caption{\textbf{Qualitative examples of BGM on PIDray~\cite{zhang2023pidray} with DINO-R50~\cite{zhang2022dino}.} BGM leverages intrinsic data properties and effectively induces the detector to focus on limited discriminative low-contrast regions, addressing the domain-specific challenge of foreground-background confusion in X-ray security images.}
  \label{fig:exp_qualitativev2}
\end{figure}

\begin{table}[h]\Huge
\centering
\caption{Instance segmentation results of Cascade Mask R-CNN~\cite{cai2018cascade} on PIDray~\cite{zhang2023pidray}, we report BBox AP and Mask AP as the evaluation metrics.}
\label{tab:seg_pidray}
\resizebox{0.48\textwidth}{!}{
\begin{tabular}{l|ccc|ccc}
\toprule
\multirow{2}{*}{\textbf{Model}} & \multicolumn{3}{c|}{\textbf{BBox AP}} & \multicolumn{3}{c}{\textbf{Mask AP}} \\
\cmidrule(r){2-4} \cmidrule(l){5-7}
& AP$^{B}$ & AP$^{B}_{50}$ & AP$^{B}_{75}$ & AP$^{M}$ & AP$^{M}_{50}$ & AP$^{M}_{75}$ \\
\midrule
C-Mask-RCNN & 69.9 & 84.2 & 77.4 & 57.1 & 81.6 & 65.6 \\
\rowcolor[rgb]{0.988,0.914,0.914}
\textbf{+BGM}             & \textbf{70.9} & \textbf{85.6} & \textbf{78.4} & \textbf{57.9} & \textbf{82.5} & \textbf{66.6} \\
\rowcolor[rgb]{0.988,0.914,0.914}
\textit{\textbf{Improvement}}             & \textbf{+1.0\%} & \textbf{+1.4\%} & \textbf{+1.0\%} & \textbf{+0.8\%} & \textbf{+0.9\%} & \textbf{+1.0\%} \\
\bottomrule
\end{tabular}
}
\label{tab:instance_seg}
\end{table}

\begin{table}[htbp]\Huge
\centering
\caption{Performance on PIDray~\cite{zhang2023pidray} at additional metrics, including \textbf{AP}, \textbf{AP\textsubscript{50}} and \textbf{AP\textsubscript{75}}, showing the effectiveness and the practical value of BGM.}
\resizebox{0.95\linewidth}{!}{
\begin{tabular}{llll|ccc}
\toprule
\multirow{2}{*}{\textbf{Type}} & \multirow{2}{*}{\textbf{Stage}} & \multirow{2}{*}{\textbf{Method}} & \multirow{2}{*}{\textbf{B-bone}} & \multicolumn{3}{c}{\textbf{Detection mAP}} \\
\cmidrule(l){5-7}
 &  &  &  & \textbf{AP} & \textbf{AP\textsubscript{50}} & \textbf{AP\textsubscript{75}} \\
\midrule
\multirow{9}{*}{CNN} 
 & \multirow{2}{*}{One} 
   & ATSS\cite{zhang2022dino} & R101 
   & 65.2 & 80.8 & 72.6 \\  
 &  
   & \cellcolor[rgb]{0.988,0.914,0.914}\textbf{ATSS+BGM} 
   & \cellcolor[rgb]{0.988,0.914,0.914}\textbf{R101} 
   & \cellcolor[rgb]{0.988,0.914,0.914}\textbf{66.4} 
   & \cellcolor[rgb]{0.988,0.914,0.914}\textbf{82.9} 
   & \cellcolor[rgb]{0.988,0.914,0.914}\textbf{73.6} \\    
 &  
   & \cellcolor[rgb]{0.988,0.914,0.914}{\textit{\textbf{Improvement}}} & \cellcolor[rgb]{0.988,0.914,0.914}{}
   & \cellcolor[rgb]{0.988,0.914,0.914}{\textbf{+1.2\%}} & \cellcolor[rgb]{0.988,0.914,0.914}{\textbf{+2.1\%}} & \cellcolor[rgb]{0.988,0.914,0.914}{\textbf{+1.0\%}} \\
\cmidrule(l){2-7}
 & \multirow{7}{*}{Two} 
   & SDANet\cite{wang2021towards} & R101 
   & 64.4 & 79.3 & 72.2 \\   
 &  
   & Improved\cite{zhang2023pidray} & R101 
   & 66.6 & 81.7 & 74.3 \\ 
 &  
   & C-RCNN\cite{cai2018cascade} & R101 
   & 68.0 & 82.6 & 75.5 \\
 &  
   & \cellcolor[rgb]{0.988,0.914,0.914}\textbf{C-RCNN+BGM} 
   & \cellcolor[rgb]{0.988,0.914,0.914}\textbf{R101} 
   & \cellcolor[rgb]{0.988,0.914,0.914}\textbf{69.5} 
   & \cellcolor[rgb]{0.988,0.914,0.914}\textbf{84.0} 
   & \cellcolor[rgb]{0.988,0.914,0.914}\textbf{76.8} \\ 
 &  
   & \cellcolor[rgb]{0.988,0.914,0.914}{\textit{\textbf{Improvement}}} & \cellcolor[rgb]{0.988,0.914,0.914}{}
   & \cellcolor[rgb]{0.988,0.914,0.914}{\textbf{+1.5\%}} & \cellcolor[rgb]{0.988,0.914,0.914}{\textbf{+1.4\%}} & \cellcolor[rgb]{0.988,0.914,0.914}{\textbf{+1.3\%}} \\
\cmidrule(l){2-7}
 &  
   & FDTNet\cite{zhu2024fdtnet} & X101 
   & 68.2 & - & - \\
 &  
   & C-RCNN\cite{cai2018cascade} & X101 
   & 69.6 & 83.7 & 77.0 \\
 &  
   & \cellcolor[rgb]{0.988,0.914,0.914}\textbf{C-RCNN+BGM} 
   & \cellcolor[rgb]{0.988,0.914,0.914}\textbf{X101} 
   & \cellcolor[rgb]{0.988,0.914,0.914}\textbf{70.6} 
   & \cellcolor[rgb]{0.988,0.914,0.914}\textbf{85.5} 
   & \cellcolor[rgb]{0.988,0.914,0.914}\textbf{78.1} \\
 &  
   & \cellcolor[rgb]{0.988,0.914,0.914}{\textit{\textbf{Improvement}}} & \cellcolor[rgb]{0.988,0.914,0.914}{}
   & \cellcolor[rgb]{0.988,0.914,0.914}{\textbf{+1.0\%}} & \cellcolor[rgb]{0.988,0.914,0.914}{\textbf{+1.8\%}} & \cellcolor[rgb]{0.988,0.914,0.914}{\textbf{+1.1\%}} \\
\midrule
\multirow{4}{*}{Transformer} 
 &  
   & DINO\cite{zhang2022dino} & R50 
   & 68.4 & 81.7 & 73.9 \\
 &  
   & \cellcolor[rgb]{0.988,0.914,0.914}\textbf{DINO+BGM} 
   & \cellcolor[rgb]{0.988,0.914,0.914}\textbf{R50} 
   & \cellcolor[rgb]{0.988,0.914,0.914}\textbf{70.1} 
   & \cellcolor[rgb]{0.988,0.914,0.914}\textbf{83.0} 
   & \cellcolor[rgb]{0.988,0.914,0.914}\textbf{75.7} \\
 &  
   & \cellcolor[rgb]{0.988,0.914,0.914}{\textit{\textbf{Improvement}}} & \cellcolor[rgb]{0.988,0.914,0.914}{}
   & \cellcolor[rgb]{0.988,0.914,0.914}{\textbf{+1.6\%}} & \cellcolor[rgb]{0.988,0.914,0.914}{\textbf{+1.3\%}} & \cellcolor[rgb]{0.988,0.914,0.914}{\textbf{+1.8\%}} \\
 &  
   & DINO\cite{zhang2022dino} & Swin 
   & 76.1 & 88.6 & 81.8 \\
 &  
   & \cellcolor[rgb]{0.988,0.914,0.914}\textbf{DINO+BGM} 
   & \cellcolor[rgb]{0.988,0.914,0.914}\textbf{Swin} 
   & \cellcolor[rgb]{0.988,0.914,0.914}\textbf{77.4} 
   & \cellcolor[rgb]{0.988,0.914,0.914}\textbf{89.8} 
   & \cellcolor[rgb]{0.988,0.914,0.914}\textbf{82.8} \\
 &  
   & \cellcolor[rgb]{0.988,0.914,0.914}{\textit{\textbf{Improvement}}} & \cellcolor[rgb]{0.988,0.914,0.914}{}
   & \cellcolor[rgb]{0.988,0.914,0.914}{\textbf{+1.3\%}} & \cellcolor[rgb]{0.988,0.914,0.914}{\textbf{+1.2\%}} & \cellcolor[rgb]{0.988,0.914,0.914}{\textbf{+1.0\%}} \\
\bottomrule
\end{tabular}}
\label{tab:pidray_performance_ap_50_75}
\end{table}

Finally, we evaluate the compatibility of BGM with classical foreground-based augmentation methods, such as TIP~\cite{bhowmik2019good} synthetic data generation. The results demonstrate that BGM maintains strong generalization capability while being fully compatible with existing foreground augmentation techniques. As a data augmentation technology that considers the detection of prohibited items in security scenes from the perspective of background, BGM is different from the way of adding constraints in the foreground, but considers the characteristics of the data and is carefully designed for the data-driven learning paradigm.

\subsubsection{Random Color Selection \textbf{\textit{v.s.}} Look-up Table }
The pseudo-color look-up table is proprietary information of the X-ray device manufacturer and is not accessible. 
As an alternative, we approximate it by analyzing the color distribution of the training set. 
Based on our statistics, the randomization of color selection results in a slightly larger set of color candidates compared to the pseudo-color look-up table~($99.4\%$ of the total).
Furthermore, we conducted additional experiments using the statistical pseudo-color look-up table.
The findings suggest that randomizing color selection leads to a slight performance improvement. 
We hypothesize that this is because the randomization not only encompasses all colors from the look-up table but also introduces additional variation, thereby slightly increasing the complexity of the training set, which may be beneficial for feature learning.
In addition, the random color strategy has better flexibility in the case of datasets facing different devices and different scenarios.

\subsubsection{Instance Segmentation}

To further validate the effectiveness of BGM, we conduct instance segmentation~(Cascade Mask RCNN~\cite{cai2018cascade})~experiments on PIDray dataset~\cite{zhang2023pidray}. As shown in Table~\ref{tab:instance_seg}, BGM consistently improves performance, \textbf{1.4\%}$\uparrow_{AP_{50}}$(BBox) and \textbf{0.9\%}$\uparrow_{AP_{50}}$(Mask), demonstrating strong generalization capabilities on different prohibited items detection tasks.

\subsubsection{Practical Analysis}
In practical applications, different scenarios may prioritize detection results at different thresholds. From a practical deployment perspective, $AP_{50}$ is more effective. Therefore, as shown in Table~\ref{tab:pidray_performance_ap_50_75} we provide additional metrics, including $AP$, $AP_{50}$ and $AP_{75}$ to further demonstrate the effectiveness of BGM. The results indicate that our method has good practical value and significantly improves the accuracy of detection compared with strong baselines.

\subsubsection{Qualitative Analysis}
Fig.~\ref{fig:exp_qualitative_full} reports qualitative samples of DINO-R50~\cite{zhang2022dino} and our method on PIDray~\cite{zhang2023pidray}, showing the effectiveness of BGM.
In addition, Fig.~\ref{fig:exp_qualitativev2} provides an intuitive and detailed visualization showing that BGM reduces failure cases observed in the baseline. 
Moreover, the imaging difference between natural light and X-ray security images is emphasized in the figure. X-ray security images are more susceptible to incoherent low-contrast regions, resulting in the degradation of detection performance.
By incorporating patch-level texture and material information, BGM improves the detection performance and broadens its focus from highly discriminative regions to low‑contrast areas. 
This effectively mitigates the domain-specific challenges arising from the physical properties of X‑ray imaging.

\section{Conclusion}




In this paper, we firstly explore the potential of data augmentation tailored for X-ray prohibited items detection \textbf{from a background perspective}.
Based on in-depth insights from the data characteristics of X-ray security images, our work contributes a straight and effective solution with our careful and refined design at both the texture level and the material level, which effectively mitigates the domain-specific challenge of imbalanced discriminative regions arising from special occlusion.
We demonstrate that BGM consistently improves the performance of detectors on multiple X-ray security benchmarks without additional annotations or training overhead.
This work highlights the importance of background modeling in data augmentation for prohibited items detection. By aligning with the data-driven learning paradigm and leveraging domain-specific characteristics, BGM offers a simple yet effective strategy to enhance model generalization in real-world security scenarios.
Moreover, BGM is compatible with existing foreground-focused augmentation technique, achieving better detection performance.
We hope this paper encourages further research on data-centric exploration to bridge the gap between different physical imaging domains.

\bibliographystyle{IEEEtran}
\bibliography{main}












\newpage

\vfill

\end{document}